\documentclass[10pt]{article} %
\usepackage[accepted]{tmlr}

\usepackage{hyperref}
\hypersetup{
 colorlinks=true, 
 citecolor=cyan,
 linkcolor=orange,
 urlcolor=cyan}
\usepackage{url}

\usepackage{multirow}
\usepackage{makecell}
\usepackage{wrapfig}  %
\usepackage{mdwlist}  %
\usepackage{sidecap}  %
\usepackage{bbm}
\usepackage{bm,amsbsy} %
\usepackage[normalem]{ulem} %
\usepackage{subcaption}
\usepackage{tikz}
\usepackage{algorithm}
\usepackage{placeins} %

\usepackage{amsmath}
\usepackage{amssymb}
\usepackage{mathtools}
\usepackage{amsthm}

\let\AND\relax
\usepackage{algorithmic}
\usepackage{booktabs}       %
\usepackage{amsfonts}       %
\usepackage{nicefrac}       %
\usepackage{microtype}      %
\usepackage{xcolor}         %
\usepackage{multirow}
\usepackage{array}
\usepackage{graphicx}
\usepackage{standalone} 
\usepackage{adjustbox}

\usepackage[capitalize,noabbrev]{cleveref}

\usepackage[makeroom]{cancel} %

\theoremstyle{plain}

\theoremstyle{definition}

\theoremstyle{remark}
\newcommand{\punt}[1]{}

\usepackage{amsthm}
\usepackage{xcolor}
\usepackage{hyperref}

\newcommand{\trp}{{^\top}} %

\renewcommand{\eqref}[1]{eq.~\ref{eq:#1}}
\newcommand{\Nrm}{\mathcal{N}}

\newcommand{\figref}[1]{Fig.~\ref{fig:#1}}  %
\newcommand{\secref}[1]{Sec.~\ref{sec:#1}}
\newcommand{\tabref}[1]{Table ~\ref{tab:#1}}  %
\newcommand{\algoref}[1]{Algorithm~\ref{algo:#1}}  %

\newcommand{\suppsecref}[1]{Appendix Sec.~\ref{supp:#1}}  

\newcommand{\vx}{\mathbf{x}}

\newcommand{\Dat}{\mathcal{D}}

\newcommand{\vphi}{\mathbf{\ensuremath{\bm{\phi}}}}

\newcommand{\vz}{\mathbf{z}}

\newcommand{\vmu}{\mathbf{\ensuremath{\bm{\mu}}}}
\newcommand{\vtheta}{\mathbf{\ensuremath{\bm{\theta}}}}
\newcommand{\LL}{\ensuremath{\mathcal{L}}}

\newcommand{\vn}{\mathbf{n}}

\newcommand{\Em}{\mathbb{E}}

\def\argmin{\mathop{\rm arg\,min}}

\def\Pr{\ensuremath{\text{Pr}}}

\newcommand{\enc}{\mathrm{Enc}}
\newcommand{\dec}{\mathrm{Dec}}

\title{DP-LDMs: Differentially Private Latent Diffusion Models}

\author{\name Michael Liu\thanks{Equal contribution.} \email mfliu@cs.ubc.ca \\
      \addr Department of Computer Science \\
      University of British Columbia \\
      \AND \\
\name Saiyue Lyu$^*$ \email saiyuel@cs.ubc.ca \\
      \addr Department of Computer Science \\
      University of British Columbia \\
      \AND \\
      \name Margarita Vinaroz$^*$\thanks{This work was done during a research visit to the Department of Computer Science at the University of British Columbia.} \email margarita.vinaroz@tuebingen.mpg.de \\
      \addr  University of Tübingen\\
    International Max Planck Research School for Intelligent Systems (IMPRS-IS) \\
      \AND \\
      \name Mijung Park \thanks{Some part of this project was done at the Technical University of Denmark.} \email mijungp@cs.ubc.ca \\
      \addr Department of Computer Science \\
      University of British Columbia
      }

\def\month{11}  %
\def\year{2024} %
\def\openreview{\url{https://openreview.net/forum?id=AkdQ266kHj}} %

\begin{document}

\maketitle

\begin{abstract}
Diffusion models (DMs) are one of the most widely used generative models for producing high quality images. However, a flurry of recent papers points out that DMs are least private forms of image generators,  by extracting a significant number of near-identical replicas of training images from DMs. 
Existing privacy-enhancing techniques for DMs, unfortunately, do not provide a good privacy-utility tradeoff. 
In this paper, we aim to improve the current state of DMs with differential privacy (DP) by adopting the \textit{Latent} Diffusion Models (LDMs). LDMs are equipped with powerful pre-trained autoencoders that map the high-dimensional pixels into lower-dimensional latent representations, in which DMs are trained, yielding a more efficient and fast training of DMs. 
Rather than fine-tuning the entire LDMs, we fine-tune only the \textit{attention} modules of LDMs with DP-SGD, reducing the number of trainable parameters by roughly $90\%$ and achieving  a better privacy-accuracy  trade-off. 
Our approach allows us to generate 
realistic, high-dimensional images (256x256) conditioned on text prompts with DP guarantees. 
Our approach provides a promising direction for training more powerful, yet training-efficient differentially private DMs, producing high-quality DP images.
Our code is available at \url{https://github.com/ParkLabML/DP-LDM}.
\end{abstract}

\section{Introduction}

A flurry of recent work 
highlights the tension between increasingly powerful diffusion models and data privacy, e.g., 
\citep{wu2023membership, carlini2023extracting, tang2023membership, hu2023membership, Duan23membershipInferenceAttacks, Matsumoto23MembershipInference, Somepalli_2023_CVPR} among many.
These papers commonly conclude that diffusion models are extremely good at memorizing training data, leaking more than twice as much training data as GANs \citep{carlini2023extracting}. This raises the question of how diffusion models should be responsibly deployed. 
\citet{carlini2023extracting} seek to use differential privacy as a remedy for the memorization problem, but their attempt was unsuccessful due to the scalability issues of differential privacy. Our paper provides a practical and scalable fine-tuning routine for diffusion models with differential privacy guarantees to avoid generating identical images to those in the private dataset.

To tackle the privacy concerns, \citet{nvidia_DPDM} propose to use DP-SGD \citep{DP_SGD} when training DMs, creating a method called \textit{DPDM}. However, DP trained DMs yield rather underwhelming performance when evaluated on datasets such as CIFAR10 and CelebA.
Recently, \citet{deepmind_DPDM} proposed a method, which is often referred to as \textit{DP-Diffusion}, to pre-train a large diffusion-based generator using public data, then fine-tune it with private data for a relatively small number of epochs using DP-SGD. 

In this paper, our goal is to further improve the performance of DP-fine-tuned DMs. To achieve this, we build on \textit{latent diffusion models (LDMs)} \citep{LDM}, which uses a pre-trained autoencoder to map the high-dimensional pixels to the so-called \textit{latent variables}, which enter into the diffusion model. The latent diffusion model defined on the lower-dimensional latent space has a significantly lower number of parameters to fine-tune than the diffusion model defined on the pixel space. 

Rather than fine-tuning the entire LDM, in our method called \textit{differentially private latent diffusion models} (DP-LDMs), we choose to fine-tune only the  \textit{attention modules} (and \textit{conditioning embedders} for conditional generation) in the LDM using DP-SGD. 
As a result, the number of trainable parameters under our approach is only 
$10\%$
of that of the diffusion models used in DP-Diffusion \citep{deepmind_DPDM} and achieves a better privacy-accuracy trade-off. 
Our choice -- fine-tune the attention modules only -- is inspired by recent observations that altering attention modules in large language models (LLMs)  substantially alters the models' behaviors 
\citep{shi2023toast, hu2021lora}.
Therefore, either manipulating or fine-tuning attention modules can yield a more targeted generation, e.g., targeted for a user-preference \citep{pasta} and  transferring to a target distribution \citep{transferLDM}. See \secref{method} for further discussion. 

The combination of considering LDMs and fine-tuning attention modules using DP-SGD is simple, \textit{yet} a solid tool whose potential impact is substantial for the following reasons:
\begin{itemize}
    \item 
    \textbf{Improved performance:} \textit{DP-LDMs outperform} many state-of-the-art methods in FID \citep{heusel2017gans} and downstream classification accuracy when evaluated on several commonly used image benchmark datasets in DP literature. This is due to unique aspects of our proposed method -- training DMs in the latent space and fine-tuning only a few selected parameters. This makes our training process considerably more efficient than training a DM from scratch with DP-SGD in DPDM \citep{nvidia_DPDM}, or fine-tuning the entire DM with DP-SGD in DP-Diffusion \citep{deepmind_DPDM}.

    \item \textbf{Significantly Reduced GPU hours:}
        Reducing the fine-tuning space in DP-LDMs not only improves the performance but also helps democratize DP image generation using DMs, which otherwise have to rely on massive computational resources only available to a small fraction of the field and would leave a huge carbon footprint. 
For instance, to generate the CIFAR10 synthetic images using an NVIDIA V100 32GB, a recent work called \textit{DP-API} by \citet{dp-api} requires 500 GPU hours and DP-Diffusion requires 1250 GPU hours (See Figure 42 in \citep{dp-api}). In our case, when using an NVIDIA RTX A4000 16GB GPU (slower than V100 32GB),  fine-tuning took 15 GPU hours, and pre-training took 192 GPU hours. Pre-training is not always necessary; we used a single pre-trained LDM for CelebA32, CIFAR10 and Camelyon17.

    \item \textbf{For-the-first-time DP Text-to-Image Generation:} \textit{We push the boundary of what DP-SGD fine-tuned generative models can achieve}, by being the first to produce high-dimensional images (256x256) at a reasonable privacy level. We showcase this in text-conditioned and class-conditioned image generation, where we input a certain text prompt (or a class label) and generate a corresponding image from a DP-fine-tuned LDM for CelebAHQ.
    These conditional, high-dimensional image generation tasks present more complex and more realistic benchmarks compared to the conventional CIFAR10 and MNIST datasets. These latter datasets, though widely used in DP image generation literature for years, are now rather simplistic and outdated. 
     Our work contributes to narrowing down the large gap between the current state of synthetic image generation in non-DP settings and that in DP settings.

     \item \textbf{Application of DP-LoRA to LDMs:} We apply the \textit{low-rank} approximation \citep{hu2021lora} to the attention modules in DMs
      to further decrease the number of parameters to fine-tune. Interestingly, the performance from LoRA was slightly worse than that of fine-tuning entire attention modules. This improvement is due to the large batch size we use, which enhances the signal-to-noise ratio and significantly boosts the performance of the fully fine-tuned model. This phenomenon is similar to what \citet{li2021large} observed when fine-tuning LLMs.
\end{itemize}

In the following section, we provide relevant background information. We then present our method along with related work and experiments on benchmark datasets.

\section{Background}

We first describe latent diffusion models (LDMs), then the definition of differential privacy (DP) and finally the DP-SGD algorithm, which we will use to train the LDMs in our method.

\subsection{Latent Diffusion Models (LDMs)}
Diffusion Models gradually denoise a normally distributed variable through learning the reverse direction of a Markov Chain of length $T$. Latent diffusion models (LDMs) by \citet{LDM} are a modification of denoising diffusion probabilistic models (DDPMs) by \citet{DDPM} in the following way. 
 First, \citet{LDM} uses a powerful auto-encoder, consisting of an encoder denoted by $\enc$ and a decoder denoted by $\dec$ . The encoder transforms a high-dimensional pixel representation $\vx$ into a lower-dimensional latent representation $\vz$ via $\vz = \enc(\vx)$; and the decoder transforms the lower-dimensional latent representation back to the original space 
 via $\hat\vx = \dec(\vz)$. 
\citet{LDM} use a combination of a perceptual loss and a patch-based adversarial objective, with extra regularization for better-controlled variance in the learned latent space, to obtain powerful autoencoders (See section 3 in \citep{LDM} for details). This training loss helps the latent representations to carry equivalent information (e.g., the spatial structure of pixels) as the pixel representations, although the dimensionality of the former is greatly reduced. 

Second, equipped with the powerful auto-encoder, \citet{LDM} trains a diffusion model (typically a UNet \citep{ronneberger2015u}) in the latent  space. Training a DM in this space can significantly expedite the training process of diffusion models, e.g., from hundreds of GPU \textit{days} to several GPU \textit{hours} for similar accuracy. 

Third, LDMs also contain \textit{attention modules} \citep{vaswani2017attention} that take inputs from a \textit{conditioning embedder}, inserted into the layers of the underlying UNet backbone as the way illustrated in \figref{schematic} to achieve flexible conditional image generation (e.g., generating images conditioning on text, image layout, class labels, etc.). 
The modified UNet is then used as a function approximator $\tau_\vtheta$ to predict an initial noise from the noisy lower-dimensional latent representations at several finite time steps $t$, where in LDMs, the noisy representations (rather than data) follow the diffusion process defined in \citet{DDPM}.

The parameters of the approximator are denoted by $\vtheta = [\vtheta^{U}, \vtheta^{Attn}, \vtheta^{Cn}]$, where $\vtheta^{U}$ are the parameters of the underlying UNet backbone excluding the parameter of attention modules $\vtheta^{Attn}$, and $\vtheta^{Cn}$ are the parameters of the conditioning embedder (We will explain these further in \secref{method}). These parameters are then optimized by minimizing the prediction error defined by 
\begin{equation}\label{eq:ldm_orig}
\LL_{ldm}(\vtheta) = \Em_{(\vz_t, y), \tau, t} \left[ \|\tau - \tau_{\vtheta}(\vz_t, t, y) \|_2^2 \right],
\end{equation} 
where $\tau \sim \Nrm(0,I)$, $t$ uniformly sampled from $\{1,\cdots,T\}$, $\vx_t$ is the noisy version of the input $\vx$ at step $t$, $\vz_t = \enc(\vx_t)$ and $y$ is what the model is conditioning on to generate data, e.g., class labels, or a prompt. 
The function approximator $\tau_{\vtheta}(\vz_t, t, y)$ takes the condition $y$, time step $t$, and latent input $\vz_t$ and then maps to an initial noise estimate.
Once the approximator is trained, the drawn samples in latent space, $\tilde{\vz}$, are transformed back to pixel space through the decoder, i.e.,  $\tilde{\vx} = \dec(\tilde{\vz})$. 
Our work introduced in \secref{method} pre-trains both auto-encoder and  $\tau_\vtheta$ using public data, then fine-tunes only $\vtheta_{Attn}, \vtheta_{Cn}$, the parameters of the attention modules and the conditioning embedded,  for private data, using DP-SGD to be explained next.

\subsection{Differential Privacy (DP)}

A mechanism $\mathcal{M}$ is  ($\epsilon$, $\delta$)-DP  for a given $\epsilon \geq 0$ and $ 0 \leq \delta < 1$ if and only if
$
\Pr[\mathcal{M}(\Dat) \in S] \leq e^\epsilon \cdot \Pr[\mathcal{M}(\Dat') \in S] + \delta
$ %
 for all possible sets of the mechanism's outputs $S$ and all neighbouring datasets $\Dat$, $\Dat'$ that differ by a single entry. A single entry difference could come from either replacing or including/excluding one entry to/from the dataset $\Dat$.

One of the most well-known and widely used DP mechanisms is the \textit{Gaussian mechanism}.
The Gaussian mechanism adds a calibrated level of noise to a function $\mu: \Dat \mapsto \mathbb{R}^p$ to ensure that the output of the mechanism is ($\epsilon, \delta$)-DP:
$\widetilde{\mu}(\Dat) = \mu(\Dat) + n$,
where $n\sim \Nrm(0, \sigma^2 \Delta_\mu^2\mathbf{I}_p)$.
Here, 
$\sigma$ is often called a privacy parameter, which is a function of $\epsilon$ and $\delta$. 
 $\Delta_\mu$ is often called the 
 {\it{global sensitivity}}
\citep{dwork2006our, dwork2014algorithmic}, which is the maximum difference in $L_2$-norm given two neighbouring $\Dat$ and $\Dat'$,  $||\mu(\Dat)-\mu(\Dat') ||_2$.
Because we are adding noise, the natural consequence is that the released function $\tilde\mu (\Dat)$ is less accurate than the non-DP counterpart, $\mu(\Dat)$. This introduces privacy-accuracy trade-offs.

DP-SGD \citep{DP_SGD} is an instantiation of the Gaussian mechanism in stochastic gradient descent (SGD) by adding an appropriate amount of Gaussian noise to the gradients in every training step, to ensure the parameters of a neural network are differentially private.
When using DP-SGD, due to the composability property of DP \cite{dwork2006our, dwork2014algorithmic}, 
privacy loss is accumulating over a typically long course of training. 
\citet{DP_SGD} exploit the subsampled Gaussian mechanism (i.e., applying the Gaussian mechanism on randomly subsampled data) to achieve a tight privacy bound. 
The \textit{Opacus} package \citep{opacus} implements the privacy analysis in DP-SGD, which we adopt in our method. One thing to note is that we use the {\bf inclusion/exclusion} definition of DP in the experiments as in \textit{Opacus}.

\section{Differentially private latent diffusion models (DP-LDMs)}\label{sec:method}

\begin{figure}[t]
    \centering
    \includegraphics[width=0.9\textwidth]{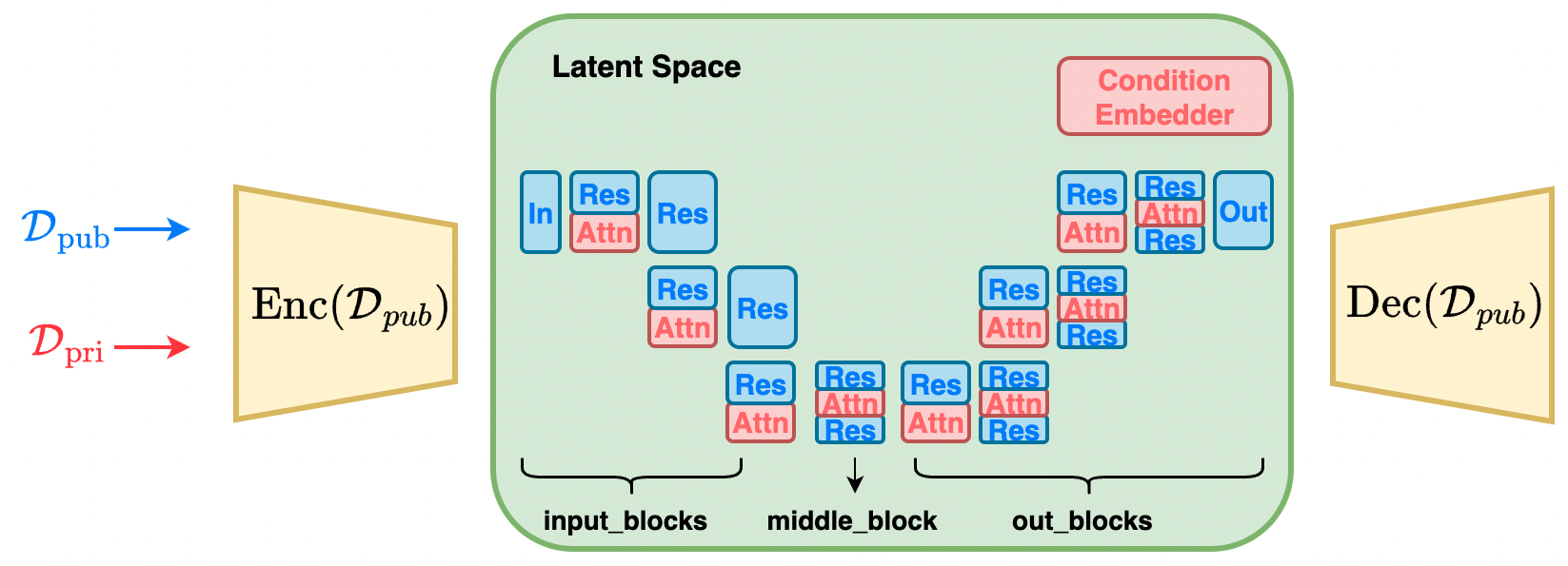}
    \caption{A schematic of DP-LDM. In the non-private step, we pre-train the auto-encoder depicted in yellow (Right and Left) with public data. We then forward pass the public data through the encoder (blue arrow on the left) to obtain latent representations. We then train the diffusion model (depicted in the green box) on the lower-dimensional latent representations. The diffusion model consists of the UNet backbone and added attention modules (in Red) with a conditioning embedder (in Red, at top-right corner). In the private step, we forward pass the private data (red arrow on the left) through the encoder to obtain latent representations of the private data. We then fine-tune only the red blocks, which are attention modules and conditioning embedder, with DP-SGD. Once the training is done, we sample the latent representations from the diffusion model, and pass them through the decoder to obtain the samples in the pixel space. }
    \label{fig:schematic}
\end{figure}

In our method, which we call \textit{differentially private latent diffusion models (DP-LDMs)}, we carry out two training steps: non-private and private steps.

\paragraph{Non-Private Step: Pre-training an autoencoder and a DM using public data.}

Following \citet{LDM}, we first pre-train an auto-encoder. The encoder scales down an image $\vx \in \mathbb{R}^{H\times W \times 3}$ to a 3-dimensional latent representation $\vz \in \mathbb{R}^{h\times w \times c}$ by a factor of $f$, where $f = H/h = W/w$. This 3-dimensional latent representation is chosen to take advantage of image-specific inductive biases that the UNet contains, e.g., 2D convolutional layers. Following \citet{LDM}, we also train the autoencoder by minimizing a combination of different losses, such as perceptual loss and adversarial loss, with some form of regularization. See \suppsecref{choose_of_autoencoder} for details.   
As noted by \citet{LDM}, we also observe that a mild form of downsampling performs the best, achieving a good balance between training efficiency and perceptually decent results.  See \suppsecref{choose_of_autoencoder} for details on different scaling factors $f=2^m$, with a different value of $m$.
Training an auto-encoder does not incur any privacy loss, as we use public data $\Dat_{pub}$ considered to be similar to private data $\Dat_{priv}$ at hand. The trained autoencoder is, therefore, a function of public data: an encoder $\enc(\Dat_{pub})$ and a decoder $\dec(\Dat_{pub})$. 

A forward pass through the trained encoder $\enc(\Dat_{pub})$ gives us a latent representation of each image, which we use to train a diffusion model.
As in \citep{LDM}, we consider a modified UNet for the function approximator $\tau_\vtheta$ shown in \figref{schematic}. 
We minimize the loss given in \eqref{ldm_orig} to estimate the parameters of $\tau_\vtheta$ as:
\begin{align}\label{eq:argmin}
\vtheta^{U}_{\Dat_{pub}},\vtheta^{Attn}_{\Dat_{pub}}, \vtheta^{Cn}_{\Dat_{pub}}
 &= \argmin_{\vtheta} \; \LL_{ldm}(\vtheta).
\end{align} Since we use public data, there is no privacy loss incurred in estimating the parameters, which are a function of public data $\Dat_{pub}$.

\paragraph{Private Step: Fine-tuning attention modules \& conditioning embedder for private data.}\label{subsec:fine-tune}

\begin{figure}[ht]%
    \centering
    \subfloat[\centering]{{\includegraphics[width=0.23\textwidth]{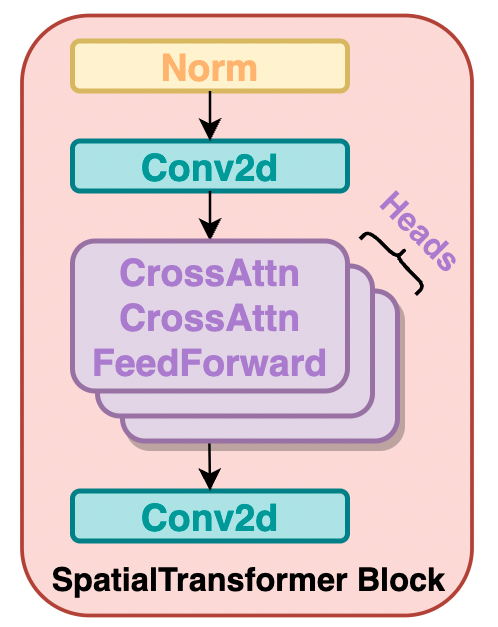} }}%
    \qquad
    \subfloat[\centering]{{\includegraphics[width=0.23\textwidth]{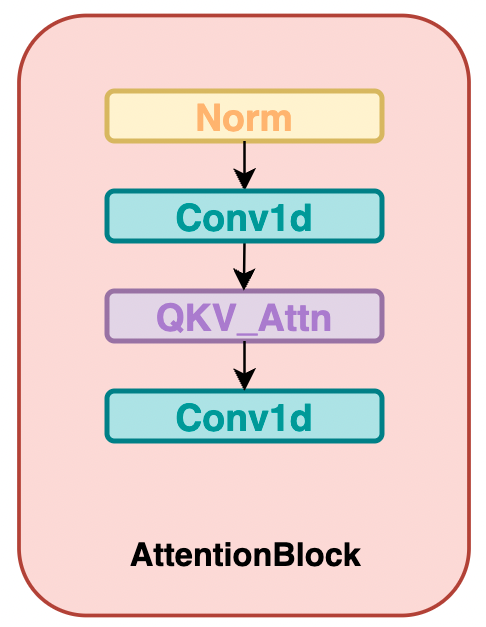} }}%
    \caption{(a) SpatialTransformer Block; (b) AttentionBlock}%
    \label{fig:block}%
\end{figure}

Given a pre-trained diffusion model, we fine-tune the attention modules and a conditioning embedder using our private data.  
For the models with the conditioned generation, the attention modules refer to  the spatial transformer blocks shown in \figref{block}(a) which contains cross-attentions and multiple heads. For the models with an unconditional generation, the attention modules refer to the attention blocks shown in \figref{block}(b). 
Consequently, 
the parameters of the attention modules, denoted by $\vtheta^{Attn}$, differ, depending on the conditioned or unconditioned cases. 
The conditioning embedder only exists in the conditioned case. Depending on the different modalities the model is trained on, the conditioning embedder takes a different form. For instance, if the model generates images conditioning on the class labels, the conditioning embedder is simply a class embedder, which embeds class labels to a latent dimension. If the model conditions on language prompts, the embedder can be a transformer.

The core part of the spatial transformer block  and the attention block is the attention layer, which has the following parameterization. Here, we explain it under the conditioned case:
\begin{align}\label{eq:Attention}
&\text{Attention}(\psi_i(\vz_t), \phi(y); Q,K,V) = \text{softmax}\left(\tfrac{QK^T}{\sqrt{d_k}}\right)V \; \in \mathbb{R}^{N \times d_k},
\end{align}
where $\psi_i(\vz_t) \in \mathbb{R}^{N \times d^{i}}$ is an intermediate representation of the latent representation $\vz_t$ through the $i$th residual convolutional block in the backbone UNet, and $\phi(y) \in \mathbb{R}^{M \times d_c}$ is the embedding of what the generation is conditioned on (e.g., class labels, or CLIP embedding). Furthermore, 
\begin{align}\label{eq:QKV}
Q=  \psi_i(\vz_t) W_Q^{(i)}\trp, \; 
K=  \phi(y) W_K^{(i)}\trp,  \; 
V= \phi(y)  W_V^{(i)} \trp, 
\end{align} 
where the parameters are denoted by 
$W_Q^{(i)} \in \mathbb{R}^{d_k \times d^{i}}$; 
$W_K^{(i)} \in \mathbb{R}^{d_k \times d_c}$; and 
$W_V^{(i)} \in \mathbb{R}^{d_k \times d_c}$. 
Unlike the conditioned case, where the key ($K$) and value ($V$) vectors are computed as a projection of the conditioning embedder, the key and value vectors are a projection of the pixel embedding $\psi_i(\vz_t)$ only in the case of the unconditioned model. 
We run DP-SGD to fine-tune these parameters to obtain differentially private $\vtheta^{Attn}_{\Dat_{priv}}$ and $\vtheta^{Cn}_{\Dat_{priv}}$,
starting from $\vtheta^{Attn}_{\Dat_{pub}}, \vtheta^{Cn}_{\Dat_{pub}}$. Our algorithm is given in \algoref{DP-LDM}.

\paragraph{Why fine-tune attention modules only?}

The output of the attention in \eqref{Attention} assigns a high focus to the features that are more important, by zooming into what truly matters in an image depending on a particular context, e.g., relevant to what the image is conditioned on. This can be quite different when we move from one distribution to the other.  By fine-tuning the attention modules (together with the conditioning embedder when conditioned case), we effectively transfer what we learned from the public data distribution to the private data distribution. However, if we fine-tune other parts of the model, e.g., the ResBlocks, the fine-tuning of these blocks can make a large change in the features themselves, degrading the performance. See \secref{experiment}. 
The idea of fine-tuning attention blocks is explored elsewhere. In fine-tuning large language models, existing work introduces a few new parameters to transformer attention blocks, and those new parameters are fine-tuned \citep{yu2022differentially, hu2021lora} to adapt to new distributions. In the context of pre-trained diffusion models, 
adding, modifying, and controlling attention layers are gaining popularity for tasks such as image editing and text-to-image generation \citep{hertz2022prompttoprompt, park2023shapeguided, zhang2023adding, transferLDM}.

\paragraph{Which public dataset to use given a private dataset?}
This is an open question in transfer learning literature. Generally, if the two datasets are close to each other in \textit{some} sense, they are assumed to be a better pair. 
However, similarity \textit{in what sense} has to be 
chosen differently depending on a particular data domain and appropriately privatized if private data is used in this step. 
For instance, if the data is in 2D space, similarity in L2-distance sense might make sense. Our data lies in a high dimensional pixel space, where judging similarity in the FID sense is widely used (when comparing two image distributions using each distribution's image samples). Hence, we use FID as a proxy to judge the similarity between two \textit{image} datasets (See \secref{simpletransfer} how we privatize this step).
In other datasets, out of the image domain, there could be a more appropriate metric to use than FID, e.g., in the case of discrete data, kernel-based distance metrics with an appropriately chosen kernel could be more useful.
\begin{algorithm}[t]
\caption{DP-LDM}
\label{algo:DP-LDM}
\begin{algorithmic}
  \STATE {\bfseries Input:} Latent representations and conditions if conditional: $\{ (\vz_{i}, y_{i}) \}_{i=1}^{N}$,
  a pre-trained model $\vtheta$, number of iterations $P$, mini-batch size $B$, clipping norm $C$, learning rate $\eta$, privacy parameter $\sigma$ corresponding to ($\epsilon$, $\delta$)-DP. Denote $\hat\vtheta = \{\vtheta^{Attn}, \vtheta^{Cn}\}$
  \FOR{$p=1$ {\bfseries to} $P$}
  \STATE  \textbf{Step 1}. Take a mini-batch $B_p$ uniformly at random with a sampling probability, $q = B/N$
  \STATE \textbf{Step 2}. For each sample $i \in B_p$ compute the gradient:\\
  \hspace{0.5cm}$ g_{p}(\vz_i, y_i) = \nabla_{\hat\vtheta_p} \LL_{ldm} (\hat\vtheta_p, \vz_i, y_i) $
  \STATE  \textbf{Step 3}. Clip the gradient: \\
  \hspace{0.5cm}$ \hat{g}_{p}(\vz_{i}, y_i) = g_{p}(\vz_{i}, y_i) / \max(1, \| g_{p}(\vz_i, y_{i}) \|_2 / C) $
  \STATE  \textbf{Step 4}. Add noise: \\
  \hspace{0.5cm}$\tilde{g}_{p} = \frac{1}{B} \Big(\sum_{i=1}^{B} \hat{g}_{p}(\vz_i, y_i) + \mathcal{N}(0,\sigma^2 C^2 I)\Big)$
  \STATE  \textbf{Step 5}. Gradient descent: $\hat\vtheta_{p+1} = \hat\vtheta_{p} - \eta \tilde{g}_p$
  \ENDFOR
  \STATE {\bfseries Return:} ($\epsilon$, $\delta$)-differentially private $\hat\vtheta_P = \{\vtheta^{Attn}_P, \vtheta^{Cn}_P\}$
\end{algorithmic}
\end{algorithm}

\section{Related Work}

Initial efforts on generating high-dimensional image data with differential privacy have focused on leveraging advanced generative models to achieve better differentially private synthetic data \citep{hu2023sok}. Some of them \citep{DPGAN, DP_CGAN, lorenzo2019, PATE_GAN, gs-wgan} utilize generative adversarial networks (GANS) \citep{GAN}, or trained GANs with the PATE structure \citep{papernot:private-training}. Other works have employed variational autoencoders (VAEs) \citep{acs2018differentially, jiang2022dp, pfitzner2022dpd}, or proposed customized structures \citep{dpmerf, dphp, dp_sinkhorn, liew2022pearl, DP-MEPF}. For instance, \citet{DP-MEPF} pretrained perceptual features using public data and privatized only data-dependent terms using maximum mean discrepancy. 

Limited works have so far delved into privatizing diffusion models. \citet{nvidia_DPDM} develop a DP score-based generative models \citep{songscore} using DP-SGD, applied to relatively simple datasets such as MNIST, FashionMNIST and CelebA (downsampled to 32$\times$32). 
\citet{deepmind_DPDM} fine-tune the ImageNet pre-trained diffusion model (DDPM) \citep{DDPM} with more than 80 M parameters using DP-SGD for CIFAR-10. We instead adopt a different model (LDM) and fine-tune only the small part of the DM in our model to achieve better privacy-accuracy trade-offs. 
As concurrent work to ours, \citet{dp-api} propose a DP-histogram mechanism to generate synthetic data 
through the utilization of publicly accessible APIs.%
Another concurrent work to ours, DP-Promise \citep{dp_promise} leverages the diffusion model's noise during the forward process to improve the privacy-accuracy trade-offs in diffusion model training.

\section{Experiments}\label{sec:experiment}

We demonstrate the performance of our method in comparison with the state-of-the-art methods in DP data generation, using several combinations of public/private data of different levels of complexity at varying privacy levels. 

{\bf Implementation.} We implemented DP-LDMs in PyTorch Lightning \citep{paszke2019pytorch} building on the LDM codebase by \citet{LDM} and Opacus \citep{opacus} for DP-SGD training. Several recent papers present the importance of using large batches in DP-SGD training to improve accuracy at a fixed privacy level \citep{how2dpfy, unlocking, bu2022scalable}. To incorporate this finding in our work, we wrote custom batch splitting code that integrates with Opacus and Pytorch Lightning, allowing us to test arbitrary batch sizes. Our DP-LDM also improves significantly with large batches as will be shown soon, consistent with the findings in recent work. For our experiments incorporating LoRA, we use the loralib \citep{hu2021lora} Python library.

{\bf Datasets\footnote{Dataset licenses: MNIST: CC BY-SA 3.0; CelebA: see \url{https://mmlab.ie.cuhk.edu.hk/projects/CelebA.html}; CIFAR-10: MIT; Camelyon17: CC0} and Evaluation.} We list all the private and public dataset pairs with corresponding evaluation metrics in \tabref{pri-pub}. Regarding high-quality generation, we use CelebAHQ for class conditional generation and Multi-Modal-CelebAHQ (MM-CelebAHQ) for text-to-image generation. Our choice of evaluation metric is either based on standard practice or following previous work to do a fair comparison. We measure the model performance by computing the Fréchet Inception Distance (FID) \citep{heusel2017gans} between the generated samples and the real data. For downstream task, we consider CNN \citep{lecun2015deep}, ResNet-9 \citep{he2016deep}, and WRN40-4 \citep{wideresnet}  to evaluate the classification performance of synthetic data.
Each number in our tables represents an average value across three independent runs, with a standard deviation (unless stated otherwise). Note that some standard deviation values are reported as $0.0$ due to rounding. Values for comparison methods are taken from their papers, with an exception for the DP-MEPF comparison to CelebAHQ, which we ran their code by loading this data. 

\begin{table}[t]
\vspace{-0.1cm}
\centering
\begin{tabular}{|l |l | c | c |}
\hline
\multirow{2}{*}{Private Dataset} & \multirow{2}{*}{Public Dataset} & Similarity & Downstream \\
  &  & Evaluation & Classifiers \\
\hline
MNIST & EMNIST(letter) & \multirow{2}{*}{-} & CNN, \\
\citep{lecun2010mnist} & \citep{cohen2017emnist} & & WRN40-4\\
\hline
CIFAR-10 & ImageNet32 & \multirow{2}{*}{FID} & ResNet-9,\\
\citep{krizhevsky2009learning} & \citep{deng2009imagenet} & & WRN40-4\\
\hline
Camelyon17-WILDS & ImageNet32 & \multirow{2}{*}{-} & \multirow{2}{*}{WRN40-4} \\
\citep{wilds} &\citep{deng2009imagenet} & & \\
\hline
CelebA32 & ImageNet32 & \multirow{2}{*}{FID} & \multirow{2}{*}{-} \\
\citep{liu2015faceattributes} &\citep{deng2009imagenet} & &\\
\hline
CelebA64 & ImageNet64 & \multirow{2}{*}{FID} & \multirow{2}{*}{ResNet-9} \\
\citep{liu2015faceattributes} & \citep{deng2009imagenet} & &\\
\hline
CelebAHQ & ImageNet256 & \multirow{2}{*}{FID} & \multirow{2}{*}{-}\\
\citep{karras2018progressive} & \citep{deng2009imagenet}& & \\
\hline
MM-CelebAHQ & LAION-400M & \multirow{2}{*}{FID} & \multirow{2}{*}{-}\\
\citep{xia2021tedigan} & \citep{schuhmann2021laion}& & \\
\hline
\end{tabular}
\caption{Private and public dataset pairs, with corresponding evaluation metric and choices of classifiers.}\label{tab:pri-pub}
\vspace{-0.1cm}
\end{table}

\subsection{Comparisons to State-of-the-art methods }\label{sec:simpletransfer}

We start with testing our method on private and public dataset pairs at varying complexity, which are generally considered to be relatively similar to each other. In particular, we present the results of transferring from Imagenet to CIFAR-10 and CelebA distributions and from EMNIST to MNIST distribution.
Additionally, to test our method's effectiveness at transferring knowledge across a large domain gap, we present the results of transferring from Imagenet to Camelyon17-WILDS. Further details on these experiments are available in appendix \ref{supp:additional_expts}.

\begin{table}[t]
\vspace{-0.1cm}
    \centering
    \begin{tabular}{|l|l| c c c c|}
         \hline
         & & $\epsilon=10$  & $\epsilon=5$ &  $\epsilon=1$ & $\epsilon =0.67$\\
         \hline
         & \textbf{DP-LDM(ours)} & \textbf{8.4 $\pm$ 0.2} & \textbf{13.4 $\pm$ 0.4}   & \textbf{22.9 $\pm$ 0.5} & \\
         CIFAR-10 & DP-Diffusion & 9.8 & 15.1 &25.2  & \\
         32x32 & DP-MEPF $(\vphi_1,\vphi_2)$  & 29.1 &  30.0 & 54.0 & \\
         $(\delta=10^{-5})$& DP-MEPF ($\vphi_1$)         &   30.3 &   35.6   &    56.5 & \\
         & DPDM            & 97.7           & -           & -         &  \\
         &DP-API & -& -& -&\textbf{7.87}\\
         &DP-Promise & 17.9& 18.9& 21.8& \\
         \hline
         & \textbf{DP-LDM(ours)} &  16.2 $\pm$ 0.2 & \textbf16.8 $\pm$ 0.3 & 25.8 $\pm$ 0.9 &\\
         CelebA& DP-MEPF $(\vphi_1)$             & 16.3           & 17.2           & 17.2  &\\
        32x32 & DP-GAN (pre-trained)            & 58.1           & 66.9           & 81.3          & \\
         $(\delta=10^{-6})$& DPDM            & 21.2           & -           & 71.8           &\\
         & DP Sinkhorn   & 189.5          &  -             &  -            & \\
         &DP-Promise & \textbf{6.0}& \textbf{6.5}&\textbf{9.0} & \\
         \hline
         & \textbf{DP-LDM(ours)} & \textbf{14.3 $\pm$ 0.1} & \textbf{16.1 $\pm$ 0.2} & 21.1 $\pm$ 0.2 &\\
         CelebA& DP-MEPF $(\vphi_1)$             & 17.4           & 16.5           & \textbf{20.4}  &\\
        64x64 & DP-GAN (pre-trained)            & 57.1           & 62.3           & 72.5          & \\
         $(\delta=10^{-6})$& DPDM            & 78.3           & -           & - & \\
         &DP-Promise & 25.3 & 26.2& 29.1& \\
         \hline
    \end{tabular}
    \caption{FID scores (lower is better) for synthetic CIFAR-10, CelebA32, and CelebA64 data, in comparison with DP-Diffusion \citep{deepmind_DPDM}, DP-MEPF \citep{DP-MEPF}, DPDM \citep{nvidia_DPDM}, DP-GAN \citep{DPGAN}, DP Sinkhorn \citep{dp_sinkhorn}, and DP-Promise \citep{dp_promise}}
    \label{tab:cifar-celeba-fid}
\end{table}

{\bf FID.} Comparison to other SOTA methods in terms of FID (the lower the better) is illustrated in \tabref{cifar-celeba-fid}. 
When tested on CIFAR-10, our DP-LDM outperforms other methods at all epsilon levels ($\epsilon = 1, 5, 10$ and $\delta=10^{-5}$) except for DP-API. Our FID values correspond to the case where only 9-16 attention modules are fine-tuned (i.e., fine-tuning only $10\%$ of trainable parameters in the model) and the rest remain fixed. See \tabref{cifar_finetuning_layers} for ablation experiments for fine-tuning different attention modules. 
When tested on CelebA32, DP-Promise achieved the best FID at all $\epsilon$ levels by a large margin compared to other existing methods. 
When tested on CelebA64,  
our unconditional LDM achieves new SOTA results at $\epsilon=10,5$ and are comparable to DP-MEPF at $\epsilon=1$. Samples are available in \textcolor{red}{\figref{celeba64_samples}}.

\begin{table}[h]
    \centering
    \begin{tabular}{|l|l| c c c|}
         \hline
         & & $\epsilon=10$  & $\epsilon=5$ &  $\epsilon=1$\\
         \hline
                                & \textbf{DP-LDM(ours)} & \textbf{65.3 $\pm$ 0.3} & \textbf{59.1 $\pm$ 0.2} & \textbf{51.3 $\pm$ 0.1}\\
         CIFAR-10               & DP-MEPF $(\vphi_1,\vphi_2)$ & 48.9 & 47.9 & 28.9 \\
         {\scriptsize ResNet-9} & DP-MEPF ($\vphi_1$)         & 51.0 & 48.5 & 29.4 \\
                                & DP-MERF                     & 13.2 & 13.4 & 13.8 \\
         \hline
         CIFAR-10              & \textbf{DP-LDM(ours)} & \textbf{78.6 $\pm$ 0.3} & - & - \\
         {\scriptsize WRN-40-4} & DP-Diffusion          & 75.6                    & - & - \\
         \hline
         \hline
         CelebA64               & \textbf{DP-LDM(ours)} & \textbf{96.4 $\pm$ 0.0} & \textbf{96.0 $\pm$ 0.0} & \textbf{94.5 $\pm$ 0.0} \\
         {\scriptsize ResNet-9} & DP-MEPF $(\vphi_1)$   & 93.9 $\pm$ 2.1          & 93.7                    & 82.9                    \\
         \hline
         \hline
         MNIST             & \textbf{DP-LDM(ours)} & 97.4 $\pm$ 0.1 & \textbf{96.8} & \textbf{95.9 $\pm$ 0.1} \\
         {\scriptsize CNN} & DPDM                  & \textbf{98.1}  & -             &   95.2                  \\
         \hline
         MNIST                 & \textbf{DP-LDM(ours)} & 97.5 $\pm$ 0.0 & - & - \\
         {\scriptsize WRN-40-4} & DP-Diffusion          & \textbf{98.6}  & - & - \\
         \hline
         \hline
         Camelyon17-    & \textbf{DP-LDM(ours)} & 85.4 $\pm$ 0.0 & - & - \\
         WILDS & DP-Diffusion          & \textbf{91.1}  & - & - \\
         {\scriptsize WRN-40-4}& DP-API & 80.5 & & \\
         \hline
    \end{tabular}
    \caption{Classification accuracies (higher is better) evaluated on real test data, when the classifiers are trained with synthetic CIFAR-10, CelebA64, and MNIST datasets. Comparison methods include DP-MEPF \citep{DP-MEPF}, DP-MERF \citep{dpmerf}, DP-Diffusion \citep{deepmind_DPDM}, and DPDM \citep{nvidia_DPDM}.  For Camelyon17 dataset, following \cite{deepmind_DPDM} and \cite{dp-api}, we set $\delta=3\cdot10^{-6}$. Each classifier's architecture is written below each name of data. We choose to use these classifiers by following previous works \citep{DP-MEPF, deepmind_DPDM, nvidia_DPDM}.}
    \label{tab:cifar-celeba-acc}
\end{table}

{\bf Downstream Classification.} FID can be viewed as a fidelity metric, serving as a proxy for the utility of the synthetic data. To directly present the utility results of the model, we also consider accuracy on the classification task, which is listed in \tabref{cifar-celeba-acc}. All the classifiers are trained with 50K synthetic samples and then evaluated on real data samples. For each dataset, we follow previous work to choose classifier models for a fair comparison. 

When tested on CIFAR-10, DP-LDM again outperforms others except DP-API. 
Compared to DP-LDM, the performance of DP-API seems more susceptible to 
the amount of domain shift from public to private data distributions. When there is a small domain gap, e.g., from Imagenet (public data) to CIFAR10 (private data), DP-API performs better than DP-LDM. However,   when there is a large domain shift between public data (Imagenet) to private data (Camelyon17),  the accuracies of downstream classifier (WRN-40-4) trained with each synthetic dataset are 85.7 for DP-LDMs and 80.5 for DP-API, respectively, at $\epsilon=10$. 
Hence, DP-LDMs seem better suited than DP-API when the domain shift is large
as fine-tuning a small part of diffusion models in DP-LDMs helps incorporate such a shift effectively.

For testing on CelebA64, we began with an LDM pre-trained on conditional ImageNet at the same resolution, and then fine-tuned it on CelebA where the (binary) class labels were given by the “Male” attribute. Our method achieves a new SOTA performance at all epsilon levels. 

When tested on MNIST, we surpass the previous methods at $\epsilon=1$ and achieve comparable results at $\epsilon=10$. 
One thing to note is that DPDM takes 1.75M parameters and 192 GPU hours to achieve 98.1 accuracy, and DP-Diffusion takes 4.2M parameters (GPU hours not showing), while our methods takes only 0.8M parameters and 10 GPU hours to achieve 97.4 accuracy, which significantly reduces the parameters and saves much computing resources. 

\begin{figure}[h]
\centering
\setlength{\tabcolsep}{1pt} %
\begin{tabular}{ccc}
\toprule
      & \small{"This woman has brown hair, and}  &  \small{"This man has sideburns,}\\
      & \small{black hair. She is attractive} &  \small{and mustache."}\\
       & \small{and wears lipstick"}   &    \\
\midrule
input  & \adjustbox{valign=m,vspace=1pt}{\includegraphics[scale=0.063]{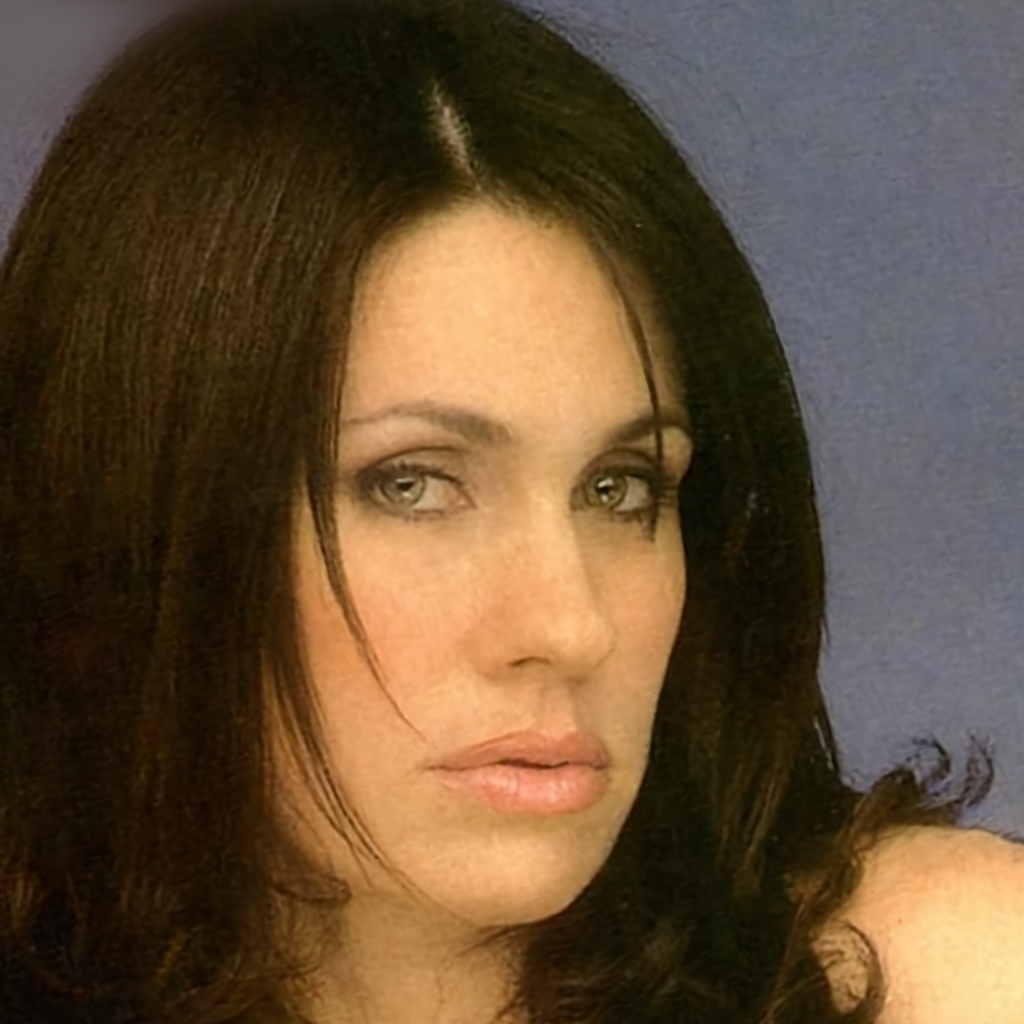}} & \adjustbox{valign=m,vspace=1pt}{\includegraphics[scale=0.063]{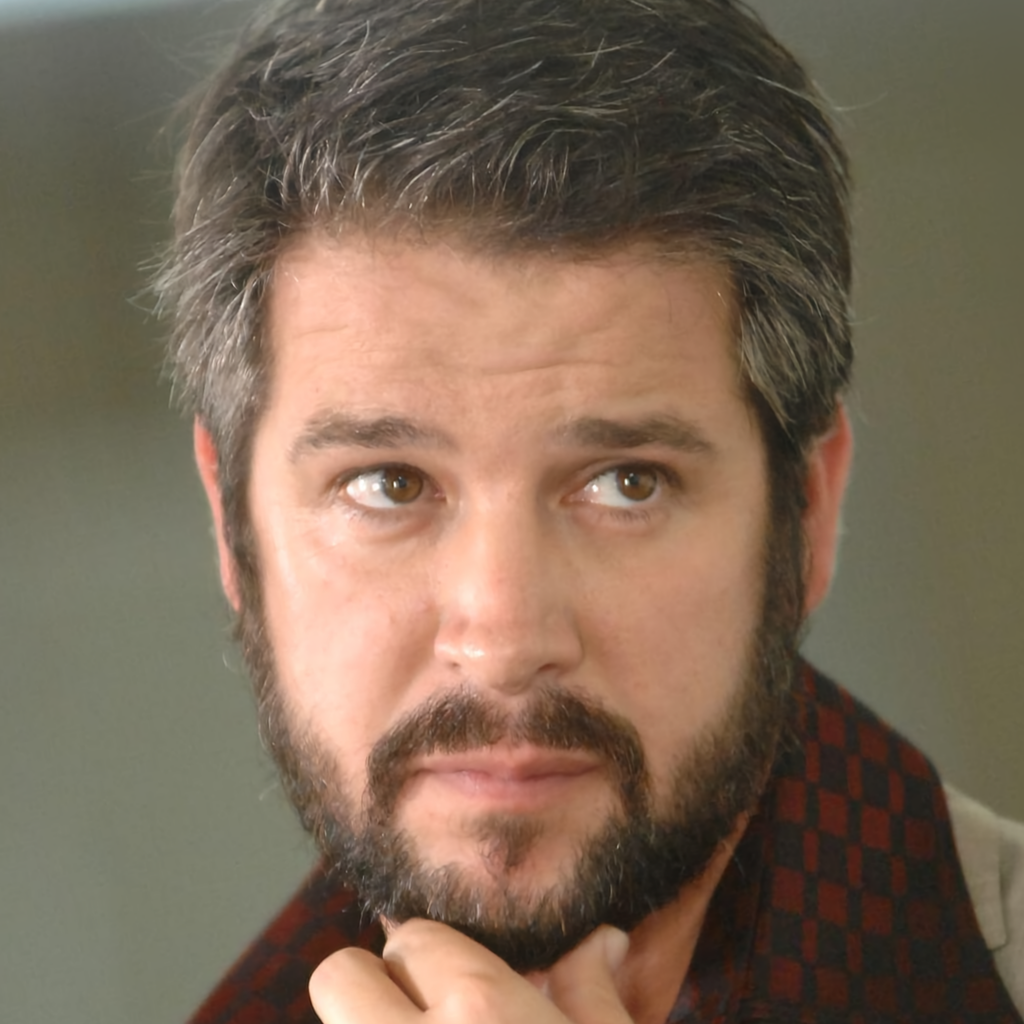}} \\
\multirow{2}{*}{$\epsilon=10$} & \adjustbox{valign=m,vspace=0pt}{\includegraphics[scale=0.25]{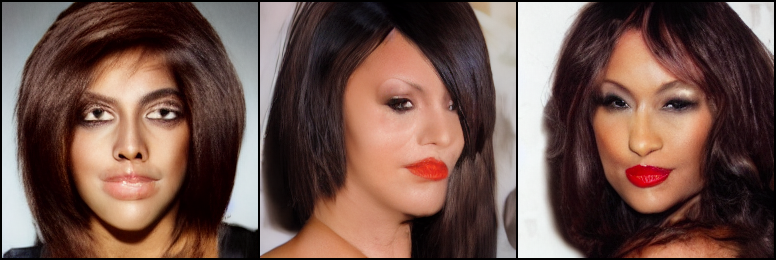}}  & \adjustbox{valign=m,vspace=0pt}{\includegraphics[scale=0.25]{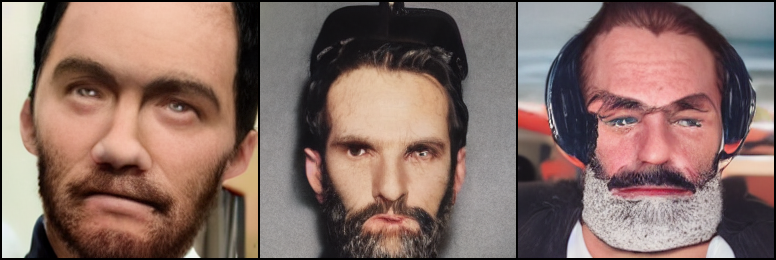}} \\
 & \adjustbox{valign=m,vspace=1pt}{\includegraphics[scale=0.25]{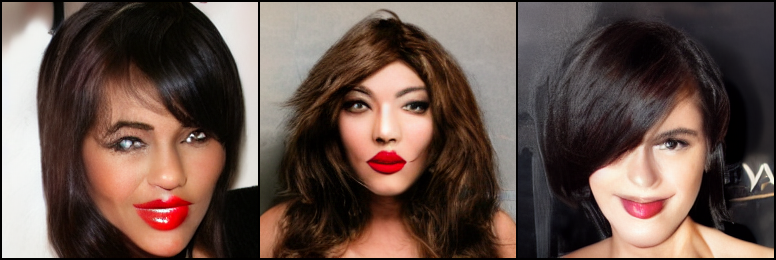}}  & \adjustbox{valign=m,vspace=1pt}{\includegraphics[scale=0.25]{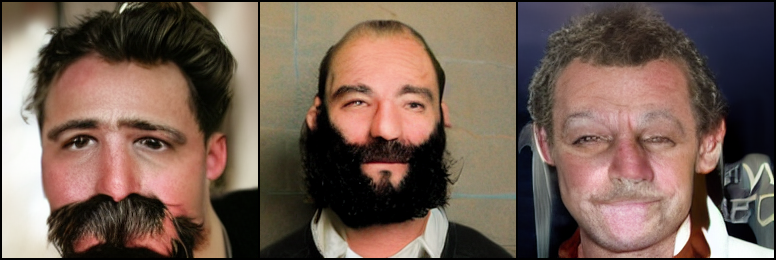}} \\
\bottomrule
\end{tabular}
\caption{Text-to-image generation of $256 \times 256$ CelebAHQ with prompts at $\epsilon=10$. FID: 15.6}
\label{fig:txt2img-celebahq}
\end{figure}

When tested on Camelyon17, our method  seems to underperform  DP-Diffusion \citep{deepmind_DPDM}. However, this could be due to their use of a pre-trained WRN-40-4 classifier (pre-trained with ImageNet32).
\citet{deepmind_DPDM} has not published their code and the pre-trained classifier (using ImageNet32) that they started fine-tuning with, so we could not directly compare the results between ours and theirs. 

\textbf{Why do we select EMNIST as a public dataset?} Previous work \citep{DP-MEPF} used SVHN as a public dataset to MNIST since they are both number images. However, SVHN and MNIST differ significantly (SVHN contains several digits per image with 3 channels while MNIST contains one digit per image with 1 channel). 
So we considered other, more similar datasets such as  EMNIST and KMNIST as public dataset candidates. 
We used FID scores to judge the closeness between public and private data, using privatized FID statistics
by following the mechanisms used in \citep{park2017dp}. See \tabref{svhn-kmnist} and \suppsecref{emnist}
which verifies our choice of EMNIST.

\subsection{Differentially private high-quality image generation}

With the latent representations of the inputs, LDMs can better improve DP training. To the best of our knowledge, we are the first to achieve high-dimensional differentially private generation.

\begin{figure}[h]
\centering
\begin{tabular}{ccc}
& \textbf{DP-LDM (Ours)} & DP-MEPF \\
\ \\
\multirow{2}{*}{$\epsilon=10$} \hspace{0.5cm} & \adjustbox{valign=m,vspace=0pt}{\includegraphics[scale=0.25, clip]{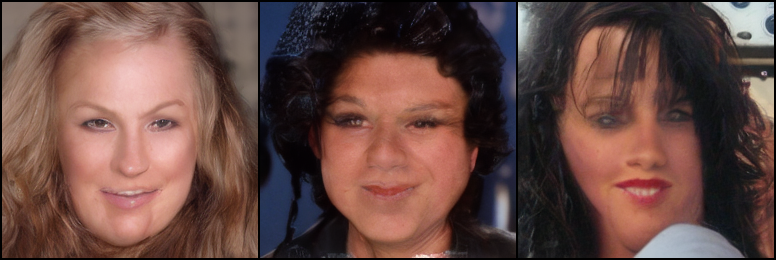}} \hspace{0.5cm} &  \adjustbox{valign=m,vspace=0pt}{\includegraphics[scale=0.25, trim=774 0 1548 258, clip]{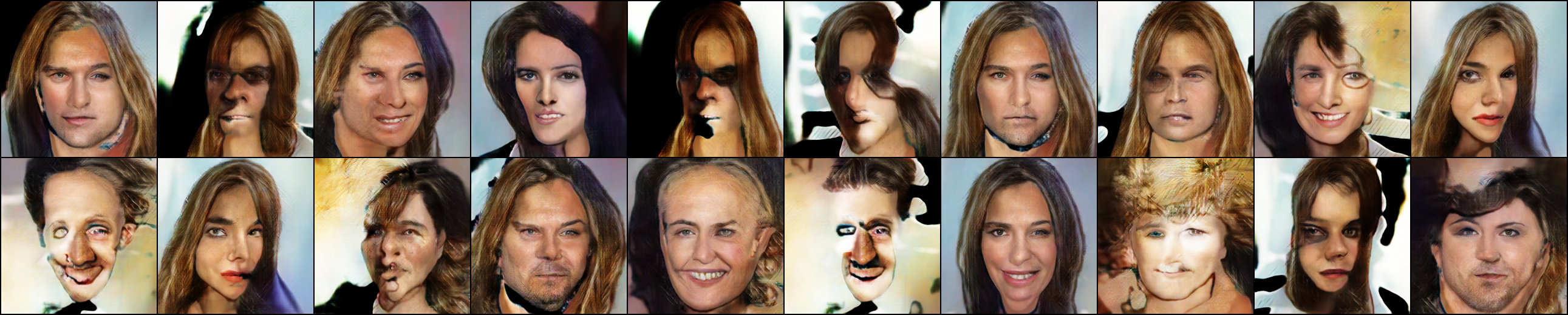}}\\
\hspace{0.5cm} & \adjustbox{valign=m,vspace=1pt}{\includegraphics[scale=0.25, clip]{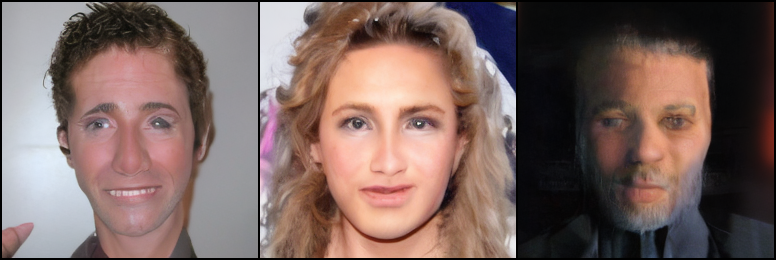}} \hspace{0.5cm} & \adjustbox{valign=m,vspace=0pt}{\includegraphics[scale=0.25, trim=1032 0 1290 258, clip]{images/celebahq/dpmepf/eps_10_dpmepf_experiment5_logs_run_18_it_45000.png}}\\
\ \\
\multirow{2}{*}{$\epsilon=1$} \hspace{0.5cm} & \adjustbox{valign=m,vspace=0pt}{\includegraphics[scale=0.25, clip]{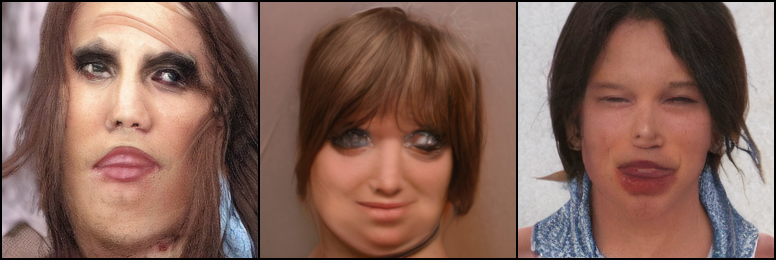}} \hspace{0.5cm} & \adjustbox{valign=m,vspace=0pt}{\includegraphics[scale=0.25, trim=1032 0 1290 258, clip]{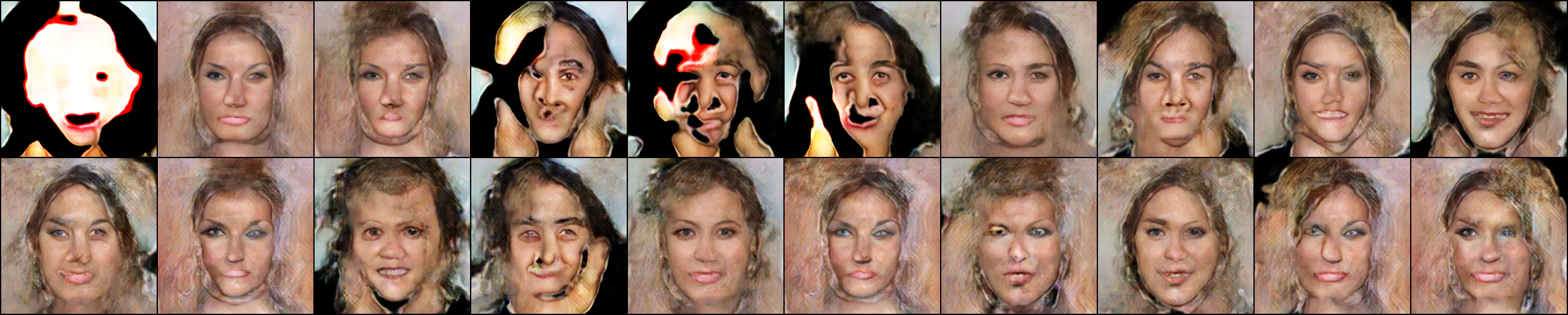}}\\
\hspace{0.5cm} & \adjustbox{valign=m,vspace=1pt}{\includegraphics[scale=0.25, clip]{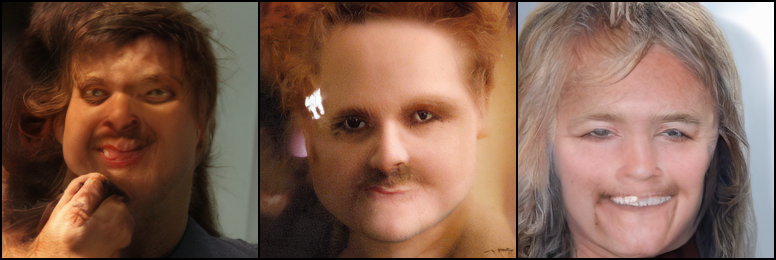}} \hspace{0.5cm} & \adjustbox{valign=m,vspace=0pt}{\includegraphics[scale=0.25, trim=1290 0 1032 258, clip]{images/celebahq/dpmepf/eps_1_dpmepf_experiment_eps1_logs_run_3_it_30000.png}}
\end{tabular}
\caption{Synthetic $256 \times 256$ CelebA samples generated at varying $\epsilon$. Samples for DP-MEPF are generated from code available in \citet{DP-MEPF}. We computed FID between our generated samples and the real data and achieve FIDs of $19.0 \pm 0.0$ at $\epsilon=10$, $20.5 \pm 0.1$ at $\epsilon=5$, and $25.6 \pm 0.1$ at $\epsilon=1$. DP-MEPF achieves an FID of $41.8$ at $\epsilon=10$ and $101.5$ at $\epsilon=1$.}
\label{fig:celebahq_samples}
\end{figure}

\textbf{Text-to-image generation.} 
For text-to-image generation, we fine-tune the LDM models pretrained with LAION-400M \citep{schuhmann2021laion} for MM-CelebAHQ (256 $\times$ 256). Each image is described by a caption, which is fed to the conditioning embedder, \textit{BERT} \citep{devlin2018bert}. We freeze the BERT embedder as well during fine-tuning attention modules to reduce the trainable parameters, then we bring back BERT for sampling. DP-LDM achieves FID scores of 15.6 for $\epsilon=10$. As illustrated in \textcolor{red}{\figref{txt2img-celebahq}}, the samples are faithful to our input prompts, but not identical to the training sample, unlike the memorization behavior of the non-private Stable Diffusion \citep{carlini2023extracting}.

\textbf{Class conditional generation.} We build our model on the LDM model provided by \citet{LDM} which is pretrained on Imagenet at a resolution of $256\times256$. Following our experiments in \secref{simpletransfer}, we fine-tune all of the SpatialTransformer blocks. While CelebAHQ does not provide class labels, each image is associated with 40 binary attributes. We choose the attribute ``Male'' to act as a binary class label for each image.
Generated samples are available in \textcolor{red}{\figref{celebahq_samples}} along with FID values. 
Compared to DP-MEPF, based on the FID scores and perceptual image quality, DP-LDM is better suited for generating detailed, plausible samples from the high-resolution dataset at a wide range of privacy levels.

\subsection{Ablation studies}

In this section, we present ablation studies for strategically improving performances and  reducing parameters.

{\bf Increasing batch size.} \tabref{celeba64_bs} (Top) shows results for DP-LDM trained with CelebA64 under different training batch sizes, which provides evidence suggesting that training with larger batch sizes improves the performance of the model.

\begin{table}[t]
\centering
\begin{tabular}{|l|c|ccc|}
\hline
& & $\epsilon=10$ & $\epsilon=5$ & $\epsilon=1$ \\
\hline
Batch&\textbf{8192}  & \textbf{14.3 $\pm$ 0.1} & \textbf{16.1 $\pm$ 0.2} & \textbf{21.1 $\pm$ 0.2}\\
Size&$2048$  & $16.2 \pm 0.2$ & $17.1 \pm 0.2$ & $22.1 \pm 0.1$ \\
&$512$   & $17.2 \pm 0.1$ & $18.0 \pm 0.1$ & $22.3 \pm 0.2$ \\
\hline
\hline
Fine-tune&\textbf{Attention Modules }  & \textbf{8.4 $\pm$ 0.2} & \textbf{13.4 $\pm$ 0.4}   & \textbf{22.9 $\pm$ 0.5}\\
Different &Resblocks  &  105.5   & 127.7 &  149.7\\
Parts &Out\_blocks  &  45.8   & 48.3 &  57.2 \\
&Input\_blocks  & 54.2    & 56.9 &  70.4\\
\hline
\end{tabular}
\caption{
    \textbf{Top}. Effect of increasing batch size on FID (CelebA64). At a fixed epsilon level, larger batches improve FIDs. 
    \textbf{Bottom}. Effect of fine-tuning only the selected part of the model (CIFAR-10). At a fixed epsilon level, fine-tuning attention modules only achieves best results.  
}
\label{tab:celeba64_bs}
\end{table}

{\bf Fine-tuning attention modules at different layers.} 
To further reduce more trainable parameters, we consider fine-tuning only a portion of attention modules See \tabref{cifar_finetuning_layers} for CIFAR-10 and \tabref{emnist-ablation} for MNIST. 
See also \suppsecref{imagenet_to_cifar} and \suppsecref{emnist-abl} for details. The best results are achieved when fine-tuning the attention modules in the out\_blocks in the UNet (out\_blocks shown in \figref{schematic}), consistently throughout all datasets we tested.
If a limited privacy budget is given, we suggest fine-tuning the attention modules in the out\_blocks only to achieve better accuracies. 

{\bf Fine-tuning a different part of the Unet.} Previous results focused on fine-tuning attention modules at varying layers. 
Here, we present the performance of fine-tuning a different part of the Unet while the rest of the model is frozen. In \tabref{celeba64_bs} (Bottom), we show the FID scores evaluated on synthetic CIFAR10 images. 
The main takeaway messages are (a) fine-tuning Resblocks hurts the performance, possibly because the learned features during the pre-training stage are altered, and (b) fine-tuning
out\_blocks is more useful than input\_blocks, while the best strategy is fine-tuning the attention modules in the out\_blocks.

\begin{table}[t]
\centering
\begin{tabular}{|l|c|ccccc |}
\hline
              & \textbf{DP-LDM} & \multicolumn{5}{c| }{LoRA (differ in rank)} \\
\hline
   Rank       &               & 64   & 8    & 4    & 2    & 1    \\
   \hline
$\epsilon=10$ & \textbf{14.3} & 16.0 & 17.0 & 18.5 & 20.6 & 22.6 \\
$\epsilon=5$  & \textbf{16.1} & 18.2 & 17.6 & 19.4 & 21.3 & 22.7 \\
$\epsilon=1$  & \textbf{21.1} & 22.3 & 23.1 & 21.5 & 24.3 & 26.3 \\
\hline
\# Parameters & 8.0M & 1.3M & 16k & 80k & 40k & 20k\\
trainable/total & 11.03\% & 1.74\% & 0.22\% & 0.11\% & 0.06\% & 0.03\%\\
\hline
\end{tabular}
\caption{
FID scores (lower is better) for incorporating LoRA into DP-LDM with different ranks with CelebA64. Each model was trained for 70 epochs.}
\label{tab:lora}
\end{table}
\begin{table}[t]
\centering
\begin{tabular}{|c|c|c||c|c|}
\hline
\multicolumn{3}{|c||}{Camelyon17}                                                     & \multicolumn{2}{|c|}{CIFAR10}                      \\ \hline
methods                & FID            & \makecell{Classification Acc \\ (WRN-40-4)} & methods             & FID           \\ \hline
DP-LDM (layer=13-16)   & \textbf{64.86} & \textbf{85.4}                               & DP-LDM (layer=9-16) & \textbf{8.4}         \\ \hline
DP-LDM (layer=1-16)    & 69.55          & -                                           & DP-LDM (layer=1-16) & 25.8                     \\ \hline
DP-LoRA (rank=8)       & 66.41          & 82.6                                        & DP-LoRA (rank=4)    & 26.77        \\ \hline
\end{tabular}
\caption{
FID scores and Classification Accuracy of DP-LoRA trained for Camelyon17 and CIFAR10, at $\epsilon=10$. The best DP-LoRA models (the best ranks shown in parentheses) still lag the best DP-LDMs (the layer ablations shown in parentheses).}
\label{tab:lora_cc}
\vspace{-0.3cm}
\end{table}
\begin{table}[t]
\centering
\scalebox{0.65}{
\begin{tabular}{|l|c|ccccccccccccccccc|}
\hline
& $\epsilon$ & \multicolumn{17}{c| }{Layer number} \\
&  & 1 & 2 & 3 & 4 & 5 & 6 & 7 & 8 & 9 & 10 & 11 & 12 & 13 & 14 & 15 & 16 & All Params \\
\hline 
pretrained& & 18.85	&17.81	&27.98	&27.82	&44.53	&45.51	&46.46	&46.01	&46.39	&46.57	&28.68	&28.93	&30.3	&20.81	&20.23	&18.03	&136.47\\
\hline
finetuned & 10 & 6.36 & 6.60 & 13.43 & 13.47 & 26.94 & 26.95 & 26.95 & 26.92 & 26.95 & 27.02 & 13.75& 13.84& 13.73& 6.83& 6.49& 6.00& 74.15\\
DP-LDM& 5 &6.52	&6.68	&13.43	&13.46	&26.93	&26.94	&26.94	&26.91	&26.93	&26.99	&13.62	&13.67	&13.59	&6.76	&6.58	&6.33	&74.09\\
&1	&6.66	&6.73	&13.43	&13.44	&26.92	&26.92	&26.93	&26.88	&26.91	&26.95	&13.48	&13.49	&13.43	&6.65	&6.66	&6.65	&73.99\\
\hline
finetuned&10	&17.09	&19.68	&53.05	&50.69	&141.35	&145.08	&145.71	&143.37	&141.03	&133.56	&47.92	&41.20	&41.25	&15.41	&14.48	&14.49	&364.64\\
DP-LoRA&5	&6.49	&6.90	&13.54	&14.10	&28.55	&29.08	&28.42	&30.26	&29.01	&28.75	&15.72	&14.90	&16.02	&7.89	&6.35	&5.68	&79.90\\
&1	&7.14	&8.30	&17.34	&16.98	&37.38	&37.50	&37.41	&38.14	&36.50	&36.90	&15.32	&14.79	&15.93	&7.34	&6.94	&7.32	&99.61\\
\hline
\end{tabular}
}
\caption{
$F$-norms of pretrained parameters (QKV matrices) are in the top part of the table; the $\ell_2$ distances between the pretrained and finetuned QKV.weight matrices for DP-LDM (Middle); DP-LoRA (Bottom). For each layer, the $F$-norm of $\frac{\alpha}{rank}\cdot BA$ is computed, and for the final column, the $\frac{\alpha}{rank}\cdot BA$ matrices are first concatenated across all layers, then the norms are computed. Following the original code of LoRA, we set $\alpha=1$.}
\label{tab:distance-lora}
\end{table}

\subsection{DP-LoRA: Applying LoRA to LDM and fine-tuning with attention modules using DP-SGD.} 

Previous work \citep{yu2022differentially} has explored LoRA during training to reduce the fine-tuning parameters. We performed additional experiments by applying LoRA to the QKV matrices in all the attention modules for CelebA64, CIFAR10, and Camelyon17 datasets. 
{
In LoRA, each QKV matrix is reparameterized as $W_{\text{new}}=W_{\text{pretrained}}+\frac{\alpha}{\text{rank}}\cdot BA$, where $W_{\text{new}}$ is the desired result of finetuning, $W_{\text{pretrained}}$ is the pretrained weight matrix, $A$ and $B$ are the low rank matrices, and $\alpha$ is a scaling hyperparameter. Following \citet{lora}, we set $\alpha$ to be equal to the first rank we try for each experiment, which in all of our runs is $1$. In DP-LoRA, A and B are updated during fine-tuning with DP-SGD and $W_{\text{pretrained}}$ is frozen.
}
As shown in \tabref{lora} and \tabref{lora_cc},
 DP-LoRA was not particularly useful under the LDMs and our current DP-LDM still outperformed.
A possible explanation could be the phenomenon previously observed in fine-tuning LLMs \citep{li2021large}:
They reasoned that the large batch size improves the signal-to-noise ratio (shown in Fig 3 [1] of \citet{li2021large}), which significantly helps improve the performance of the model with full updates, compared to low-rank updates.

For more curious readers, we  show the empirical distances between $W_{\text{new}}$ and $W_{\text{pretrained}}$ for the LoRA models fine-tuned for CelebA64. We compute the $\ell_2$ distance between the prerained and finetuned \textit{qkv.weight} matrices for DP-LDM (vanilla), and the $F$-norm of $\frac{\alpha}{rank}\cdot BA$ for DP-LoRA, in \tabref{distance-lora}. 
Two remarks: 
First, given that the domain shift between ImageNet and CelebA is relatively large, the adaptation matrices in LoRA, in some cases (e.g., at $\epsilon=10$), completely overwhelm $W_{\text{pretrained}}$.
We suspect that the larger F. norms in the case of DP-LoRA (relative to DP-LDM) are due to its updates being restricted to be low-rank. While DP-LDM may find a configuration of parameter values that achieve low loss (and good performance) relatively closer to the initial parameters, DP-LoRA may need to search further in parameter space for a set of parameter values that achieves similar performance.
Second, 
  the fact that DP-LoRA gets the best FID score at rank 64 at $\epsilon=10$ (the smallest amount of noise we tried out) while it does at rank 4 at $\epsilon=1$ and rank 8 at $\epsilon=5$, the optimal adaptation matrix in the UNet might not be necessarily rank-deficient as in the transformer case.
Next, we visualize the fine-tuned attention modules in an attempt to gain insights into our fine-tuned models.

\begin{figure}[t]
\centering
\begin{subfigure}{.49\textwidth}
    \centering
    \includegraphics[width=1\linewidth]{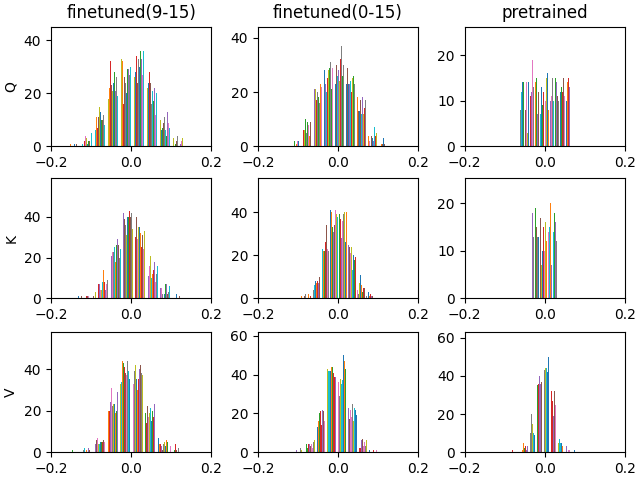}  
    \caption{Histogram of $W_q$,$W_k$,$W_v$ in each model.}
\end{subfigure}
\begin{subfigure}{.49\textwidth}
    \centering
    \includegraphics[width=1\linewidth]{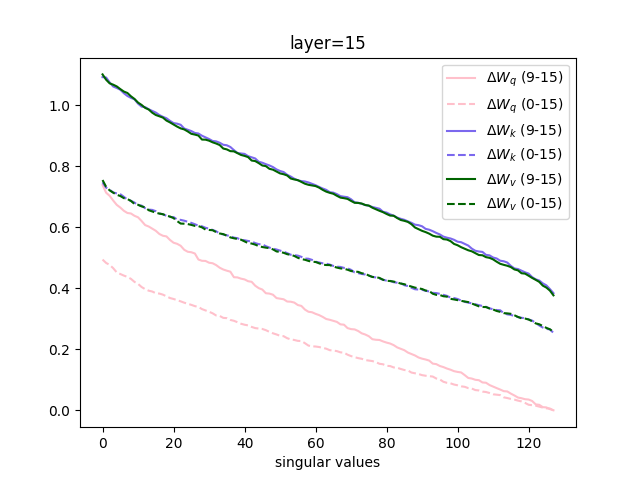}  
    \caption{Singular values of $\Delta W_q$, $\Delta W_k$, $\Delta W_v$ in each model.}
\end{subfigure}
\caption{The Q,K,V matrices in cross-attention modules in two fine-tuned models for CIFAR10 at $\epsilon=10$. Fine-tuned(0-15) indicates attention modules of all layers are fine-tuned, while Fine-tuned(9-15) indicates only the attention modules at layers 9-15 are fine-tuned.
(a) Each row corresponds to $W_q$(top), $W_k$(middle), and $W_v$(bottom).
(b) Pink represents $\Delta W_q$, blue $\Delta W_k$, and green $\Delta W_v$, respectively. Solid lines represent the model(9-15) and dotted lines represent the model(0-15).
}
\label{fig:CIFAR10_hist_svals}
\vspace{-0.5cm}
\end{figure}

\subsection{Visualization of fine-tuned attention modules}\label{subsec:look_at_attn}
We first show the histogram of $W_q, W_k, W_v$ matrices in cross-attention modules before and after fine-tuning for CIFAR10 in (a) of 
\figref{CIFAR10_hist_svals}.
Because we fine-tune only the attention modules while the intermediate representations are fixed, the distributions over $W_q, W_k, W_v$ matrices after fine-tuning (first and second columns) are significantly different from those before fine-tuning (third column). 
When fine-tuning a smaller number of layers (9-15), larger changes have to be made per layer (indicated by the large spread in the histogram for $W_q,W_k,W_v$ matrices) to compensate for the distributional shift from $\Dat_{pub}$ to $\Dat_{priv}$, compared to fine-tuning all layers (0-15). 

\begin{figure}[h]
\centering
\begin{subfigure}{.49\textwidth}
    \centering
\includegraphics[width=1\linewidth]{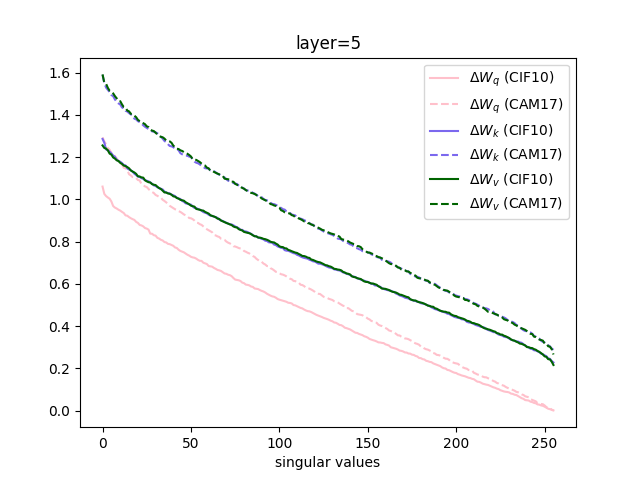}  
    \caption{Singular values of $\Delta W_q$, $\Delta W_k$, $\Delta W_v$ at layer 5}
\end{subfigure}
\begin{subfigure}{.49\textwidth}
    \centering
    \includegraphics[width=1\linewidth]{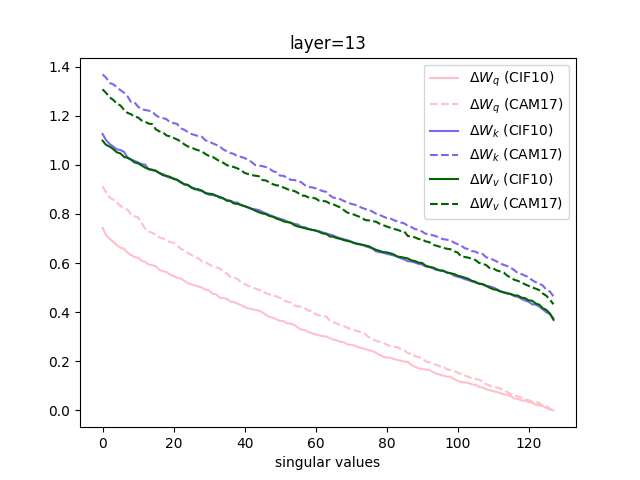}  
    \caption{Singular values of $\Delta W_q$, $\Delta W_k$, $\Delta W_v$ at layer 13}
\end{subfigure}
\caption{Singular values of $\Delta W_q$, $\Delta W_k$, $\Delta W_v$ in cross-attention modules in the conditional LDMs fine-tuned for CIFAR10 (CIF10) and that for Camelyon17 (CAM17), at $\epsilon=10$ with $\delta=10^{-5}$ for CIFAR10 and $\delta=3\cdot10^{-6}$ for Camelyon17. 
Dotted lines represent CAM17, while solid lines represent CIF10. 
Purple and green solid lines are overlapping in both plots.
}
\label{fig:CAM17_CIF10_svals}
\vspace{-0.5cm}
\end{figure}

We then computed the singular values of $\Delta W_q = W_{q,\text{finetuned}}-W_{q, \text{pretrained}}$ ($\Delta W_k, \Delta W_v$ are defined with respect to $k,v$ matrices).
As shown in (b) of 
\figref{CIFAR10_hist_svals},
$\Delta W$ for K,V matrices are correlated possibly because the role of $W_k$ and $W_v$ is to pick which class embeddings are more useful (both are multiplied by the class embedding as in  \eqref{QKV}), their fine-tuned values remain similar.
The singular values fall off fast in the case of  $\Delta W_q$, while those of $\Delta W_k$ and $\Delta W_v$ fall off slowly and even the smallest singular values are far from zero. 
Similarly, in \figref{CAM17_CIF10_svals}, 
the singular values of fine-tuned model for Camelyon17 seem to decay faster than those for CIFAR10. 
However, the smallest singular values at the later layer are again far from zero.
This might imply low rank approximation to these matrices can cause losing information. This observation is  consistent with the performance of DP-LoRA lagging that of DP-LDM shown in \tabref{lora} and \tabref{lora_cc}.

\begin{figure}[t]
\centering
\begin{subfigure}{.49\textwidth}
    \centering
\includegraphics[width=1\linewidth]{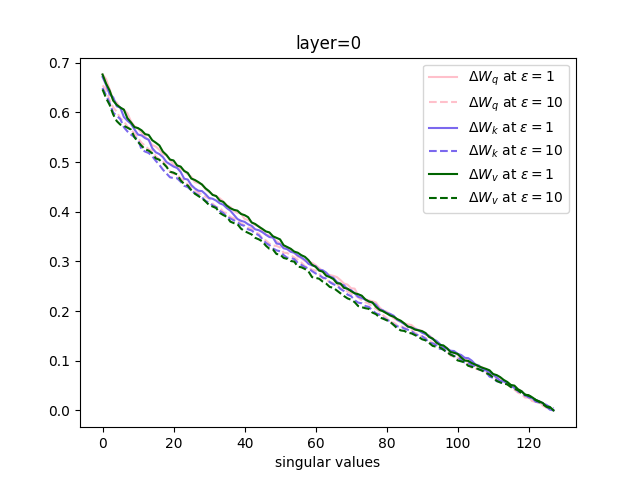}  
    \caption{Layer 0}
\end{subfigure}
\begin{subfigure}{.49\textwidth}
    \centering
    \includegraphics[width=1\linewidth]{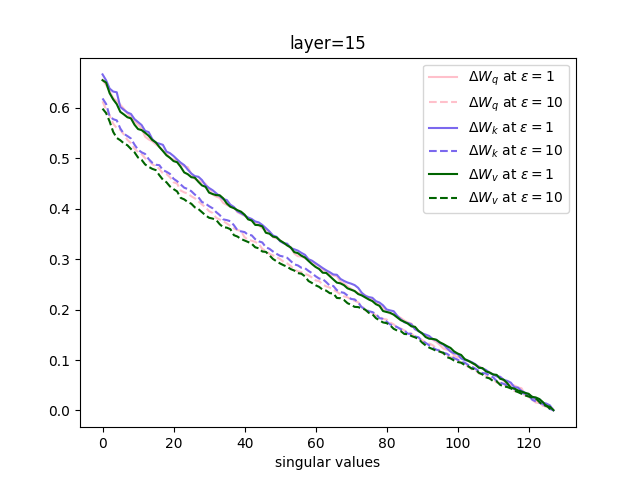}  
    \caption{Layer 15}
\end{subfigure}
\caption{Singular values of $\Delta W_q$, $\Delta W_k$, $\Delta W_v$ in self-attention modules at varying layers of an unconditional LDM, fine-tuned for CelebA64, at $\epsilon=10$ and $\epsilon=1$. 
Dotted lines represent the model fine-tuned at $\epsilon=10$, and solid lines that at $\epsilon=1$. }
\label{fig:Celeba64_svals}
\vspace{-0.5cm}
\end{figure}

\figref{Celeba64_svals} shows the singular values of $\Delta W_q$, $\Delta W_k$, $\Delta W_v$ in the unconditional LDM fine-tuned for CelebA64. 
There is only a slight difference between the singular values of the model fine-tuned at $\epsilon=1$ and the model fine-tuned at $\epsilon=10$. 
This could be explained by 
using a fixed random seed during training. 
As the random seed determines the order of training batches as well as the direction of the random noise injected by DP-SGD, each model converges to similar areas of the parameter space. Thus, differences in the fine-tuned weight matrices arise from the different magnitudes of noise added.
On the other hand, this is no longer the case, when the target datasets change like in  \figref{CAM17_CIF10_svals}, or when the ablation setting (which layers to be fine-tuned) changes like in \figref{CIFAR10_hist_svals}, even if we set the seed number to be the same across different settings.

\section{Conclusion}

In \textit{Differentially Private Latent Diffusion Models} (DP-LDM), we utilize DP-SGD to fine-tune only the attention modules (and embedders for conditioned generation) of the pretrained LDM at varying layers with privacy-sensitive data. We demonstrate that our method is capable of generating quality images in various scenarios. We perform an in-depth analysis of the ablation of DP-LDM to explore the strategy for reducing parameters for more applicable training of DP-SGD. Based on our promising results, we conclude that fine-tuning LDMs is an efficient and effective framework for DP generative learning. We hope our results can contribute to future research in DP data generation, considering the rapid advances in diffusion-based generative modelling. 

\section*{Broader Impact Statement}

As investigated in \citet{carlini2023extracting}, diffusion models can memorize individual images from the training data set and generate an identical synthetic image to the training image. Aiming to impact society positively, we provide a method to fine-tune latent diffusion models with differential privacy guarantees so that the fine-tuned models do not generate identical images to those in the private data. 

Our method relies on public data for better scalability of differential privacy, which needs some attention. As \citet{tramer2022considerations} pointed out, public data themselves may still contain sensitive information.
From our perspective and as many other DP generative modelling papers also noticed, auxiliary public data still emerges as the most promising option for attaining satisfactory utility, for the models at this large scale. 
We hope for better-curated public datasets to be available in the near future.

During the review process, one of the reviewers mentioned that generative tools are becoming more and more realistic and of general availability, with the risk of generating harmful content, especially when generating human-related content. Broadly speaking, our method does not stop malicious users from creating harmful content, as the diffusion model will generate images based on any input prompt given by the malicious user. However, what our method can do is protect the privacy of the individuals whose data was in the private dataset, meaning the faces in the private dataset will not appear when generating images from the DP fine-tuned models.

To encourage reproducible research practices, our code is available at \url{https://github.com/ParkLabML/DP-LDM} with detailed instructions. And all the hyperparameters are discussed in detail in \suppsecref{hyper}.
\section*{Limitations}

While the paper proposes a technique to counter-attack the privacy risks of diffusion models, the evaluation has not been done against existing privacy attacks. So we know that the diffusion model is private ``under the definition of DP", but 
we caution practitioners against making an assumption that a DP image generative model protects private data against all possible attack methods.

This may not be a direct limitation of our work itself, but it is restrictive that there is no agreement within the community on which definition of privacy should be used to assess the privacy of generated images.
\citet{carlini2023extracting} provided a single example of exploitation, but it remains unclear what aspects we aim to protect in generated data. Unfortunately, the privacy parameters $\epsilon$ and $\delta$ in our method are especially difficult to interpret in the image domain. While private images may be protected, it is unclear whether perceptually similar images enjoy the same level of privacy. 
Further work is required to suggest more appropriate definitions of privacy for practitioners to employ to avoid providing a vague sense of privacy in the image generation domain.

\subsubsection*{Acknowledgments}
We thank our anonymous reviewers for their constructive feedback, which has helped significantly improve
our manuscript. We thank the Digital Research Alliance of Canada (Compute Canada) for its computational
resources and services. 
M. Liu was funded by the Canada Graduate Scholarships — Master’s program of the Natural Sciences and Engineering Research Council of Canada (NSERC).
S. Lyu, M. Vinaroz, and M. Park were supported in part by the Natural Sciences and Engineering
Research Council of Canada (NSERC) and the Canada CIFAR AI Chairs program.
M. Park was also partially funded by Novo Nordisk Fonden RECUIT grant no.0065800 during her stay at the Technical University of Denmark.

\bibliography{main}

\begin{thebibliography}{68}
\providecommand{\natexlab}[1]{#1}
\providecommand{\url}[1]{\texttt{#1}}
\expandafter\ifx\csname urlstyle\endcsname\relax
  \providecommand{\doi}[1]{doi: #1}\else
  \providecommand{\doi}{doi: \begingroup \urlstyle{rm}\Url}\fi

\bibitem[Abadi et~al.(2016)Abadi, Chu, Goodfellow, McMahan, Mironov, Talwar, and Zhang]{DP_SGD}
Martin Abadi, Andy Chu, Ian~J. Goodfellow, H.~Brendan McMahan, Ilya Mironov, Kunal Talwar, and Li~Zhang.
\newblock Deep learning with differential privacy.
\newblock In \emph{Proceedings of the 2016 ACM SIGSAC Conference on Computer and Communications Security}, CCS ’16, pp.\  308–318, New York, NY, USA, 2016. Association for Computing Machinery.
\newblock ISBN 9781450341394.
\newblock \doi{10.1145/2976749.2978318}.

\bibitem[Acs et~al.(2018)Acs, Melis, Castelluccia, and De~Cristofaro]{acs2018differentially}
Gergely Acs, Luca Melis, Claude Castelluccia, and Emiliano De~Cristofaro.
\newblock Differentially private mixture of generative neural networks.
\newblock \emph{IEEE Transactions on Knowledge and Data Engineering}, 31\penalty0 (6):\penalty0 1109--1121, 2018.

\bibitem[Bie et~al.(2022)Bie, Kamath, and Zhang]{bie2022private}
Alex Bie, Gautam Kamath, and Guojun Zhang.
\newblock Private {GAN}s, revisited.
\newblock In \emph{NeurIPS 2022 Workshop on Synthetic Data for Empowering ML Research}, 2022.

\bibitem[Bu et~al.(2022)Bu, Mao, and Xu]{bu2022scalable}
Zhiqi Bu, Jialin Mao, and Shiyun Xu.
\newblock Scalable and efficient training of large convolutional neural networks with differential privacy, 2022.

\bibitem[Cao et~al.(2021)Cao, Bie, Vahdat, Fidler, and Kreis]{dp_sinkhorn}
Tianshi Cao, Alex Bie, Arash Vahdat, Sanja Fidler, and Karsten Kreis.
\newblock Don't generate me: Training differentially private generative models with sinkhorn divergence.
\newblock In \emph{Neural Information Processing Systems (NeurIPS)}, 2021.

\bibitem[Carlini et~al.(2023)Carlini, Hayes, Nasr, Jagielski, Sehwag, Tram\`{e}r, Balle, Ippolito, and Wallace]{carlini2023extracting}
Nicholas Carlini, Jamie Hayes, Milad Nasr, Matthew Jagielski, Vikash Sehwag, Florian Tram\`{e}r, Borja Balle, Daphne Ippolito, and Eric Wallace.
\newblock Extracting training data from diffusion models.
\newblock In \emph{Proceedings of the 32nd USENIX Conference on Security Symposium}, SEC '23, USA, 2023. USENIX Association.
\newblock ISBN 978-1-939133-37-3.

\bibitem[Chen et~al.(2020)Chen, Orekondy, and Fritz]{gs-wgan}
Dingfan Chen, Tribhuvanesh Orekondy, and Mario Fritz.
\newblock Gs-wgan: A gradient-sanitized approach for learning differentially private generators.
\newblock In \emph{Advances in Neural Information Processing Systems 33}, 2020.

\bibitem[Cohen et~al.(2017)Cohen, Afshar, Tapson, and Van~Schaik]{cohen2017emnist}
Gregory Cohen, Saeed Afshar, Jonathan Tapson, and Andre Van~Schaik.
\newblock Emnist: Extending mnist to handwritten letters.
\newblock In \emph{2017 international joint conference on neural networks (IJCNN)}, pp.\  2921--2926. IEEE, 2017.

\bibitem[De et~al.(2022)De, Berrada, Hayes, Smith, and Balle]{unlocking}
Soham De, Leonard Berrada, Jamie Hayes, Samuel~L. Smith, and Borja Balle.
\newblock Unlocking high-accuracy differentially private image classification through scale, 2022.

\bibitem[Deng et~al.(2009)Deng, Dong, Socher, Li, Li, and Fei-Fei]{deng2009imagenet}
Jia Deng, Wei Dong, Richard Socher, Li-Jia Li, Kai Li, and Li~Fei-Fei.
\newblock Imagenet: A large-scale hierarchical image database.
\newblock In \emph{2009 IEEE conference on computer vision and pattern recognition}, pp.\  248--255. Ieee, 2009.

\bibitem[Devlin et~al.(2018)Devlin, Chang, Lee, and Toutanova]{devlin2018bert}
Jacob Devlin, Ming-Wei Chang, Kenton Lee, and Kristina Toutanova.
\newblock Bert: Pre-training of deep bidirectional transformers for language understanding.
\newblock \emph{arXiv preprint arXiv:1810.04805}, 2018.

\bibitem[Dockhorn et~al.(2023)Dockhorn, Cao, Vahdat, and Kreis]{nvidia_DPDM}
Tim Dockhorn, Tianshi Cao, Arash Vahdat, and Karsten Kreis.
\newblock Differentially private diffusion models, 2023.
\newblock URL \url{https://openreview.net/forum?id=pX21pH4CsNB}.

\bibitem[Duan et~al.(2023)Duan, Kong, Wang, Shi, and Xu]{Duan23membershipInferenceAttacks}
Jinhao Duan, Fei Kong, Shiqi Wang, Xiaoshuang Shi, and Kaidi Xu.
\newblock Are diffusion models vulnerable to membership inference attacks?
\newblock In \emph{Proceedings of the 40th International Conference on Machine Learning}, ICML'23. JMLR.org, 2023.

\bibitem[Dwork et~al.(2006)Dwork, Kenthapadi, McSherry, Mironov, and Naor]{dwork2006our}
Cynthia Dwork, Krishnaram Kenthapadi, Frank McSherry, Ilya Mironov, and Moni Naor.
\newblock Our data, ourselves: Privacy via distributed noise generation.
\newblock In \emph{Advances in Cryptology - EUROCRYPT 2006, 25th Annual International Conference on the Theory and Applications of Cryptographic Techniques}, volume 4004 of \emph{Lecture Notes in Computer Science}, pp.\  486--503. Springer, 2006.
\newblock \doi{10.1007/11761679_29}.

\bibitem[Dwork et~al.(2014)Dwork, Roth, et~al.]{dwork2014algorithmic}
Cynthia Dwork, Aaron Roth, et~al.
\newblock The algorithmic foundations of differential privacy.
\newblock \emph{Foundations and Trends{\textregistered} in Theoretical Computer Science}, 9\penalty0 (3--4):\penalty0 211--407, 2014.

\bibitem[Frigerio et~al.(2019)Frigerio, de~Oliveira, Gomez, and Duverger]{lorenzo2019}
Lorenzo Frigerio, Anderson~Santana de~Oliveira, Laurent Gomez, and Patrick Duverger.
\newblock Differentially private generative adversarial networks for time series, continuous, and discrete open data.
\newblock In \emph{{ICT} Systems Security and Privacy Protection - 34th {IFIP} {TC} 11 International Conference, {SEC} 2019, Lisbon, Portugal, June 25-27, 2019, Proceedings}, pp.\  151--164, 2019.
\newblock \doi{10.1007/978-3-030-22312-0\_11}.

\bibitem[Ghalebikesabi et~al.(2023)Ghalebikesabi, Berrada, Gowal, Ktena, Stanforth, Hayes, De, Smith, Wiles, and Balle]{deepmind_DPDM}
Sahra Ghalebikesabi, Leonard Berrada, Sven Gowal, Ira Ktena, Robert Stanforth, Jamie Hayes, Soham De, Samuel~L. Smith, Olivia Wiles, and Borja Balle.
\newblock Differentially private diffusion models generate useful synthetic images, 2023.

\bibitem[{Goodfellow} et~al.(2014){Goodfellow}, {Pouget-Abadie}, {Mirza}, {Xu}, {Warde-Farley}, {Ozair}, {Courville}, and {Bengio}]{GAN}
I.~J. {Goodfellow}, J.~{Pouget-Abadie}, M.~{Mirza}, B.~{Xu}, D.~{Warde-Farley}, S.~{Ozair}, A.~{Courville}, and Y.~{Bengio}.
\newblock Generative adversarial networks.
\newblock In \emph{Advances in Neural Information Processing Systems}, 2014.

\bibitem[Harder et~al.(2021)Harder, Adamczewski, and Park]{dpmerf}
Frederik Harder, Kamil Adamczewski, and Mijung Park.
\newblock {DP-MERF}: Differentially private mean embeddings with random features for practical privacy-preserving data generation.
\newblock In \emph{AISTATS}, volume 130 of \emph{Proceedings of Machine Learning Research}, pp.\  1819--1827. PMLR, 2021.

\bibitem[Harder et~al.(2023)Harder, Jalali, Sutherland, and Park]{DP-MEPF}
Frederik Harder, Milad Jalali, Danica~J. Sutherland, and Mijung Park.
\newblock Pre-trained perceptual features improve differentially private image generation.
\newblock \emph{Transactions on Machine Learning Research}, 2023.
\newblock ISSN 2835-8856.
\newblock URL \url{https://openreview.net/forum?id=R6W7zkMz0P}.

\bibitem[He et~al.(2016)He, Zhang, Ren, and Sun]{he2016deep}
Kaiming He, Xiangyu Zhang, Shaoqing Ren, and Jian Sun.
\newblock Deep residual learning for image recognition.
\newblock In \emph{Proceedings of the IEEE conference on computer vision and pattern recognition}, pp.\  770--778, 2016.

\bibitem[Hertz et~al.(2022)Hertz, Mokady, Tenenbaum, Aberman, Pritch, and Cohen-Or]{hertz2022prompttoprompt}
Amir Hertz, Ron Mokady, Jay Tenenbaum, Kfir Aberman, Yael Pritch, and Daniel Cohen-Or.
\newblock Prompt-to-prompt image editing with cross attention control, 2022.

\bibitem[Heusel et~al.(2017)Heusel, Ramsauer, Unterthiner, Nessler, and Hochreiter]{heusel2017gans}
Martin Heusel, Hubert Ramsauer, Thomas Unterthiner, Bernhard Nessler, and Sepp Hochreiter.
\newblock Gans trained by a two time-scale update rule converge to a local nash equilibrium.
\newblock \emph{Advances in neural information processing systems}, 30, 2017.

\bibitem[Ho et~al.(2020)Ho, Jain, and Abbeel]{DDPM}
Jonathan Ho, Ajay Jain, and Pieter Abbeel.
\newblock Denoising diffusion probabilistic models.
\newblock In H.~Larochelle, M.~Ranzato, R.~Hadsell, M.F. Balcan, and H.~Lin (eds.), \emph{Advances in Neural Information Processing Systems}, volume~33, pp.\  6840--6851. Curran Associates, Inc., 2020.
\newblock URL \url{https://proceedings.neurips.cc/paper_files/paper/2020/file/4c5bcfec8584af0d967f1ab10179ca4b-Paper.pdf}.

\bibitem[Hu et~al.(2021)Hu, Shen, Wallis, Allen-Zhu, Li, Wang, Wang, and Chen]{hu2021lora}
Edward~J. Hu, Yelong Shen, Phillip Wallis, Zeyuan Allen-Zhu, Yuanzhi Li, Shean Wang, Lu~Wang, and Weizhu Chen.
\newblock Lora: Low-rank adaptation of large language models, 2021.

\bibitem[Hu et~al.(2022)Hu, Shen, Wallis, Allen-Zhu, Li, Wang, Wang, and Chen]{lora}
Edward~J Hu, Yelong Shen, Phillip Wallis, Zeyuan Allen-Zhu, Yuanzhi Li, Shean Wang, Lu~Wang, and Weizhu Chen.
\newblock Lo{RA}: Low-rank adaptation of large language models.
\newblock In \emph{International Conference on Learning Representations}, 2022.
\newblock URL \url{https://openreview.net/forum?id=nZeVKeeFYf9}.

\bibitem[Hu \& Pang(2023)Hu and Pang]{hu2023membership}
Hailong Hu and Jun Pang.
\newblock Membership inference of diffusion models, 2023.

\bibitem[Hu et~al.(2023)Hu, Wu, Li, Long, Garrido, Ge, Ding, Forsyth, Li, and Song]{hu2023sok}
Yuzheng Hu, Fan Wu, Qinbin Li, Yunhui Long, Gonzalo Garrido, Chang Ge, Bolin Ding, David Forsyth, Bo~Li, and Dawn Song.
\newblock Sok: Privacy-preserving data synthesis.
\newblock In \emph{2024 IEEE Symposium on Security and Privacy (SP)}, pp.\  2--2. IEEE Computer Society, 2023.

\bibitem[Jiang et~al.(2022)Jiang, Zhang, Karami, Chen, Shao, and Yu]{jiang2022dp}
Dihong Jiang, Guojun Zhang, Mahdi Karami, Xi~Chen, Yunfeng Shao, and Yaoliang Yu.
\newblock Dp$^2$-vae: Differentially private pre-trained variational autoencoders.
\newblock \emph{arXiv preprint arXiv:2208.03409}, 2022.

\bibitem[Karras et~al.(2018)Karras, Aila, Laine, and Lehtinen]{karras2018progressive}
Tero Karras, Timo Aila, Samuli Laine, and Jaakko Lehtinen.
\newblock Progressive growing of {GAN}s for improved quality, stability, and variation.
\newblock In \emph{International Conference on Learning Representations}, 2018.
\newblock URL \url{https://openreview.net/forum?id=Hk99zCeAb}.

\bibitem[Koh et~al.(2021)Koh, Sagawa, Marklund, Xie, Zhang, Balsubramani, Hu, Yasunaga, Phillips, Gao, Lee, David, Stavness, Guo, Earnshaw, Haque, Beery, Leskovec, Kundaje, Pierson, Levine, Finn, and Liang]{wilds}
Pang~Wei Koh, Shiori Sagawa, Henrik Marklund, Sang~Michael Xie, Marvin Zhang, Akshay Balsubramani, Weihua Hu, Michihiro Yasunaga, Richard~Lanas Phillips, Irena Gao, Tony Lee, Etienne David, Ian Stavness, Wei Guo, Berton~A. Earnshaw, Imran~S. Haque, Sara Beery, Jure Leskovec, Anshul Kundaje, Emma Pierson, Sergey Levine, Chelsea Finn, and Percy Liang.
\newblock Wilds: A benchmark of in-the-wild distribution shifts, 2021.

\bibitem[Krizhevsky et~al.(2009)Krizhevsky, Hinton, et~al.]{krizhevsky2009learning}
Alex Krizhevsky, Geoffrey Hinton, et~al.
\newblock Learning multiple layers of features from tiny images.
\newblock Technical report, University of Toronto, Toronto, ON, Canada, 2009.

\bibitem[LeCun \& Cortes(2010)LeCun and Cortes]{lecun2010mnist}
Yann LeCun and Corinna Cortes.
\newblock {MNIST} handwritten digit database.
\newblock http://yann.lecun.com/exdb/mnist/, 2010.
\newblock URL \url{http://yann.lecun.com/exdb/mnist/}.

\bibitem[LeCun et~al.(2015)LeCun, Bengio, and Hinton]{lecun2015deep}
Yann LeCun, Yoshua Bengio, and Geoffrey Hinton.
\newblock Deep learning.
\newblock \emph{nature}, 521\penalty0 (7553):\penalty0 436--444, 2015.

\bibitem[Li et~al.(2022)Li, Tramer, Liang, and Hashimoto]{li2021large}
Xuechen Li, Florian Tramer, Percy Liang, and Tatsunori Hashimoto.
\newblock Large language models can be strong differentially private learners.
\newblock In \emph{International Conference on Learning Representations}, 2022.

\bibitem[Liew et~al.(2022{\natexlab{a}})Liew, Takahashi, and Ueno]{liew2022pearl}
Seng~Pei Liew, Tsubasa Takahashi, and Michihiko Ueno.
\newblock {PEARL}: Data synthesis via private embeddings and adversarial reconstruction learning.
\newblock In \emph{International Conference on Learning Representations}, 2022{\natexlab{a}}.

\bibitem[Liew et~al.(2022{\natexlab{b}})Liew, Takahashi, and Ueno]{pearl}
Seng~Pei Liew, Tsubasa Takahashi, and Michihiko Ueno.
\newblock {PEARL}: Data synthesis via private embeddings and adversarial reconstruction learning.
\newblock In \emph{International Conference on Learning Representations}, 2022{\natexlab{b}}.

\bibitem[Lin et~al.(2023)Lin, Gopi, Kulkarni, Nori, and Yekhanin]{dp-api}
Zinan Lin, Sivakanth Gopi, Janardhan Kulkarni, Harsha Nori, and Sergey Yekhanin.
\newblock Differentially private synthetic data via foundation model apis 1: Images, 2023.

\bibitem[Liu et~al.(2015)Liu, Luo, Wang, and Tang]{liu2015faceattributes}
Ziwei Liu, Ping Luo, Xiaogang Wang, and Xiaoou Tang.
\newblock Deep learning face attributes in the wild.
\newblock In \emph{Proceedings of International Conference on Computer Vision (ICCV)}, December 2015.

\bibitem[Matsumoto et~al.(2023)Matsumoto, Miura, and Yanai]{Matsumoto23MembershipInference}
Tomoya Matsumoto, Takayuki Miura, and Naoto Yanai.
\newblock Membership inference attacks against diffusion models.
\newblock In \emph{2023 IEEE Security and Privacy Workshops (SPW)}, pp.\  77--83, 2023.
\newblock \doi{10.1109/SPW59333.2023.00013}.

\bibitem[Papernot et~al.(2017)Papernot, Abadi, Erlingsson, Goodfellow, and Talwar]{papernot:private-training}
Nicolas Papernot, Martín Abadi, \'{U}lfar Erlingsson, Ian~J. Goodfellow, and Kunal Talwar.
\newblock Semi-supervised knowledge transfer for deep learning from private training data.
\newblock In \emph{Proceedings of the International Conference on Learning Representations (ICLR)}, 2017.

\bibitem[Park et~al.(2023)Park, Luo, Toste, Azadi, Liu, Karalashvili, Rohrbach, and Darrell]{park2023shapeguided}
Dong~Huk Park, Grace Luo, Clayton Toste, Samaneh Azadi, Xihui Liu, Maka Karalashvili, Anna Rohrbach, and Trevor Darrell.
\newblock Shape-guided diffusion with inside-outside attention, 2023.

\bibitem[Park et~al.(2017)Park, Foulds, Choudhary, and Welling]{park2017dp}
Mijung Park, James Foulds, Kamalika Choudhary, and Max Welling.
\newblock Dp-em: Differentially private expectation maximization.
\newblock In \emph{Artificial Intelligence and Statistics}, pp.\  896--904. PMLR, 2017.

\bibitem[Paszke et~al.(2019)Paszke, Gross, Massa, Lerer, Bradbury, Chanan, Killeen, Lin, Gimelshein, Antiga, et~al.]{paszke2019pytorch}
Adam Paszke, Sam Gross, Francisco Massa, Adam Lerer, James Bradbury, Gregory Chanan, Trevor Killeen, Zeming Lin, Natalia Gimelshein, Luca Antiga, et~al.
\newblock Pytorch: An imperative style, high-performance deep learning library.
\newblock \emph{Advances in neural information processing systems}, 32, 2019.

\bibitem[Pfitzner \& Arnrich(2022)Pfitzner and Arnrich]{pfitzner2022dpd}
Bjarne Pfitzner and Bert Arnrich.
\newblock Dpd-fvae: Synthetic data generation using federated variational autoencoders with differentially-private decoder.
\newblock \emph{arXiv preprint arXiv:2211.11591}, 2022.

\bibitem[Ponomareva et~al.(2023)Ponomareva, Vassilvitskii, Xu, McMahan, Kurakin, and Zhang]{how2dpfy}
Natalia Ponomareva, Sergei Vassilvitskii, Zheng Xu, Brendan McMahan, Alexey Kurakin, and Chiyaun Zhang.
\newblock How to dp-fy ml: A practical tutorial to machine learning with differential privacy.
\newblock In \emph{Proceedings of the 29th ACM SIGKDD Conference on Knowledge Discovery and Data Mining}, KDD '23, pp.\  5823–5824, New York, NY, USA, 2023. Association for Computing Machinery.
\newblock ISBN 9798400701030.
\newblock \doi{10.1145/3580305.3599561}.
\newblock URL \url{https://doi.org/10.1145/3580305.3599561}.

\bibitem[Rombach et~al.(2022)Rombach, Blattmann, Lorenz, Esser, and Ommer]{LDM}
Robin Rombach, Andreas Blattmann, Dominik Lorenz, Patrick Esser, and Bj\"orn Ommer.
\newblock High-resolution image synthesis with latent diffusion models.
\newblock In \emph{Proceedings of the IEEE/CVF Conference on Computer Vision and Pattern Recognition (CVPR)}, pp.\  10684--10695, June 2022.

\bibitem[Ronneberger et~al.(2015)Ronneberger, Fischer, and Brox]{ronneberger2015u}
Olaf Ronneberger, Philipp Fischer, and Thomas Brox.
\newblock U-net: Convolutional networks for biomedical image segmentation.
\newblock In \emph{Medical Image Computing and Computer-Assisted Intervention--MICCAI 2015: 18th International Conference, Munich, Germany, October 5-9, 2015, Proceedings, Part III 18}, pp.\  234--241. Springer, 2015.

\bibitem[Schuhmann et~al.(2021)Schuhmann, Vencu, Beaumont, Kaczmarczyk, Mullis, Katta, Coombes, Jitsev, and Komatsuzaki]{schuhmann2021laion}
Christoph Schuhmann, Richard Vencu, Romain Beaumont, Robert Kaczmarczyk, Clayton Mullis, Aarush Katta, Theo Coombes, Jenia Jitsev, and Aran Komatsuzaki.
\newblock Laion-400m: Open dataset of clip-filtered 400 million image-text pairs.
\newblock \emph{arXiv preprint arXiv:2111.02114}, 2021.

\bibitem[Shi et~al.(2023)Shi, Gai, Darrell, and Wang]{shi2023toast}
Baifeng Shi, Siyu Gai, Trevor Darrell, and Xin Wang.
\newblock Toast: Transfer learning via attention steering, 2023.

\bibitem[Somepalli et~al.(2023)Somepalli, Singla, Goldblum, Geiping, and Goldstein]{Somepalli_2023_CVPR}
Gowthami Somepalli, Vasu Singla, Micah Goldblum, Jonas Geiping, and Tom Goldstein.
\newblock Diffusion art or digital forgery? investigating data replication in diffusion models.
\newblock In \emph{Proceedings of the IEEE/CVF Conference on Computer Vision and Pattern Recognition (CVPR)}, pp.\  6048--6058, June 2023.

\bibitem[Song et~al.(2021)Song, Sohl-Dickstein, Kingma, Kumar, Ermon, and Poole]{songscore}
Yang Song, Jascha Sohl-Dickstein, Diederik~P Kingma, Abhishek Kumar, Stefano Ermon, and Ben Poole.
\newblock Score-based generative modeling through stochastic differential equations.
\newblock \emph{International Conference on Learning Representations}, 2021.

\bibitem[Tang et~al.(2023)Tang, Wu, Aydore, Kearns, and Roth]{tang2023membership}
Shuai Tang, Zhiwei~Steven Wu, Sergul Aydore, Michael Kearns, and Aaron Roth.
\newblock Membership inference attacks on diffusion models via quantile regression, 2023.

\bibitem[Torkzadehmahani et~al.(2019)Torkzadehmahani, Kairouz, and Paten]{DP_CGAN}
Reihaneh Torkzadehmahani, Peter Kairouz, and Benedict Paten.
\newblock Dp-cgan: Differentially private synthetic data and label generation.
\newblock In \emph{The IEEE Conference on Computer Vision and Pattern Recognition (CVPR) Workshops}, June 2019.

\bibitem[Tram{\`e}r et~al.(2022)Tram{\`e}r, Kamath, and Carlini]{tramer2022considerations}
Florian Tram{\`e}r, Gautam Kamath, and Nicholas Carlini.
\newblock Considerations for differentially private learning with large-scale public pretraining.
\newblock \emph{arXiv preprint arXiv:2212.06470}, 2022.

\bibitem[Vaswani et~al.(2017)Vaswani, Shazeer, Parmar, Uszkoreit, Jones, Gomez, Kaiser, and Polosukhin]{vaswani2017attention}
Ashish Vaswani, Noam Shazeer, Niki Parmar, Jakob Uszkoreit, Llion Jones, Aidan~N Gomez, {\L}ukasz Kaiser, and Illia Polosukhin.
\newblock Attention is all you need.
\newblock \emph{Advances in neural information processing systems}, 30, 2017.

\bibitem[Vinaroz et~al.(2022)Vinaroz, Charusaie, Harder, Adamczewski, and Park]{dphp}
Margarita Vinaroz, Mohammad-Amin Charusaie, Frederik Harder, Kamil Adamczewski, and Mi~Jung Park.
\newblock Hermite polynomial features for private data generation.
\newblock In \emph{ICML}, volume 162 of \emph{Proceedings of Machine Learning Research}, pp.\  22300--22324. PMLR, 2022.

\bibitem[Wang et~al.(2024)Wang, Pang, Zhigang, Rao, Zhou, and Minhui]{dp_promise}
Haichen Wang, Shuchao Pang, Lu~Zhigang, Yihang Rao, Yongbin Zhou, and Xue Minhui.
\newblock dp-promise: Differentially private diffusion probabilistic models for image synthesis.
\newblock In \emph{33rd USENIX Security Symposium (USENIX Security 24)}, Philadelphia, PA, August 2024. USENIX Association.
\newblock URL \url{https://www.usenix.org/conference/usenixsecurity24/presentation/wang-haichen}.

\bibitem[Wu et~al.(2023)Wu, Yu, Li, Backes, and Zhang]{wu2023membership}
Yixin Wu, Ning Yu, Zheng Li, Michael Backes, and Yang Zhang.
\newblock Membership inference attacks against text-to-image generation models, 2023.
\newblock URL \url{https://openreview.net/forum?id=J41IW8Z7mE}.

\bibitem[Xia et~al.(2021)Xia, Yang, Xue, and Wu]{xia2021tedigan}
Weihao Xia, Yujiu Yang, Jing-Hao Xue, and Baoyuan Wu.
\newblock Tedigan: Text-guided diverse face image generation and manipulation.
\newblock In \emph{Proceedings of the IEEE/CVF conference on computer vision and pattern recognition}, pp.\  2256--2265, 2021.

\bibitem[Xie et~al.(2018)Xie, Lin, Wang, Wang, and Zhou]{DPGAN}
Liyang Xie, Kaixiang Lin, Shu Wang, Fei Wang, and Jiayu Zhou.
\newblock Differentially private generative adversarial network.
\newblock \emph{arXiv preprint arXiv:1802.06739}, 2018.

\bibitem[Yoon et~al.(2019)Yoon, Jordon, and van~der Schaar]{PATE_GAN}
Jinsung Yoon, James Jordon, and Mihaela van~der Schaar.
\newblock {PATE}-{GAN}: Generating synthetic data with differential privacy guarantees.
\newblock In \emph{International Conference on Learning Representations}, 2019.

\bibitem[You \& Zhao(2023)You and Zhao]{transferLDM}
Fuming You and Zhou Zhao.
\newblock Transferring pretrained diffusion probabilistic models, 2023.
\newblock URL \url{https://openreview.net/forum?id=8u9eXwu5GAb}.

\bibitem[Yousefpour et~al.(2021)Yousefpour, Shilov, Sablayrolles, Testuggine, Prasad, Malek, Nguyen, Ghosh, Bharadwaj, Zhao, Cormode, and Mironov]{opacus}
Ashkan Yousefpour, Igor Shilov, Alexandre Sablayrolles, Davide Testuggine, Karthik Prasad, Mani Malek, John Nguyen, Sayan Ghosh, Akash Bharadwaj, Jessica Zhao, Graham Cormode, and Ilya Mironov.
\newblock Opacus: {U}ser-friendly differential privacy library in {PyTorch}.
\newblock \emph{arXiv preprint arXiv:2109.12298}, 2021.

\bibitem[Yu et~al.(2022)Yu, Naik, Backurs, Gopi, Inan, Kamath, Kulkarni, Lee, Manoel, Wutschitz, Yekhanin, and Zhang]{yu2022differentially}
Da~Yu, Saurabh Naik, Arturs Backurs, Sivakanth Gopi, Huseyin~A Inan, Gautam Kamath, Janardhan Kulkarni, Yin~Tat Lee, Andre Manoel, Lukas Wutschitz, Sergey Yekhanin, and Huishuai Zhang.
\newblock Differentially private fine-tuning of language models.
\newblock In \emph{International Conference on Learning Representations}, 2022.
\newblock URL \url{https://openreview.net/forum?id=Q42f0dfjECO}.

\bibitem[Zagoruyko \& Komodakis(2017)Zagoruyko and Komodakis]{wideresnet}
Sergey Zagoruyko and Nikos Komodakis.
\newblock Wide residual networks, 2017.

\bibitem[Zhang et~al.(2023{\natexlab{a}})Zhang, Rao, and Agrawala]{zhang2023adding}
Lvmin Zhang, Anyi Rao, and Maneesh Agrawala.
\newblock Adding conditional control to text-to-image diffusion models, 2023{\natexlab{a}}.

\bibitem[Zhang et~al.(2023{\natexlab{b}})Zhang, Singh, Liu, Liu, Yu, Gao, and Zhao]{pasta}
Qingru Zhang, Chandan Singh, Liyuan Liu, Xiaodong Liu, Bin Yu, Jianfeng Gao, and Tuo Zhao.
\newblock Tell your model where to attend: Post-hoc attention steering for llms, 2023{\natexlab{b}}.

\end{thebibliography}
\bibliographystyle{tmlr}

\newpage
\appendix
\begin{center}
    {\LARGE\textbf{Appendix}}
\end{center}

\section{Additional Experiments}
\label{supp:additional_expts}

\subsection{Scaling factor effect in pre-training the autoencoder}\label{supp:choose_of_autoencoder}

In \tabref{factor_ch_effect}, we provide FID results after pre-training the autoencoder with Imagenet dataset for different scaling factors $f$ and number of channels. 

\begin{table}[ht]
\centering
\scalebox{1}{
\begin{tabular}{l | ccc}
&  \multicolumn{3}{c } {$\#$ channels } \\
&  128 & 64 & 32  \\
 \hline
$f=2$ & \textbf{27.6} & 36.4 & 46.8 \\
$f=4$ &  32.9 & 51.0 & 83.5  \\
\hline
\end{tabular}}
\caption{
FID scores (lower is better) for pre-trained autoencoders with different $f$ and number of channels.}
\label{tab:factor_ch_effect}
\end{table}

\subsection{Transferring from EMNIST to MNIST  distribution}\label{supp:emnist}

Here are the details when we compare DP-LDM to existing methods with the most common DP settings $\epsilon=1, 10$ and $\delta=10^{-5}$ in \tabref{mnist_eps}.

\begin{table*}[htb]
\centering
\scalebox{0.8}{
\begin{tabular}{llcccccc}
\toprule
& & DP-LDM (Ours)  & DP-DM &  DP-Diffusion & DP-HP & PEARL & DPGANr\\
\midrule
\multirow{2}{*}{$\epsilon=10$} & CNN & $97.4\pm$ 0.1 & \textbf{98.1}  & \textbf{-}  & \textbf{-} & 78.8 & 95.0\\
& WRN & $97.5\pm$ 0.0 & \textbf{-} & \textbf{98.6} & \textbf{-}  & \textbf{-} & \textbf{-}\\
\midrule
\multirow{1}{*}{$\epsilon=1$} & CNN & \textbf{95.9$\pm$ 0.1} & 95.2 & \textbf{-} & 81.5 & 78.2 &  80.1 \\
\midrule
& \# params & \textbf{0.8M} & 1.75M &  4.2M  & \textbf{-}  & \textbf{-} & \textbf{-} \\
& GPU Hours & \textbf{10h} & 192h &  \textbf{-}  & \textbf{-}  & \textbf{-} & \textbf{-} \\
\bottomrule
\end{tabular}}
\caption{
Downstream accuracies by CNN, MLP and WRN-40-4, evaluated on the generated MNIST data samples. We compare our results with existing work DPDM \citep{nvidia_DPDM}, DP-Diffusion \citep{deepmind_DPDM}, PEARL \citep{pearl}, DPGANr \citep{bie2022private}, and DP-HP \citep{dphp}. The GPU hours is for DP training only. The GPU hours for pretraining steps of our method are present in \tabref{autoencoder} and \tabref{dm}.}
\label{tab:mnist_eps}
\end{table*}

\subsubsection{Choosing public dataset with DP constraint}

FID scores are commonly used for measuring the similarity of two dataset. It first uses a pre-trained neural network (such as InceptionV3) to extract features from both datasets; then fits two Gaussian distributions to both datasets respectively, via computing the mean and covariance of the feature representations for both of them; then use the means and covariances to calculate the Fréchet distance. Following \citet{park2017dp}, we computed the FID scores between public data and private data in a differentially private manner. I.e. consider a data matrix by $X$, where $n$ i.i.d. observations in a privacy-sensitive dataset are stacked in each row of $X$. We denote each observation of this dataset by $\vx_i\in\mathbb{R}^d$. Hence, $X \in \mathbb{R}^{n \times d}$.
We denote the inception feature given a datapoint $\vx_i$ by $\vphi(\vx_i)$.
We further denote the mean and the covariance of the inception features, computed on a public dataset, by $\vmu_0$ and $\Sigma_0$. Similarly, we denote those computed on a privacy-sensitive dataset by $\vmu,\Sigma$. The non-DP computation of FID is given by the following formula (notations are only used in this subsection):
\begin{align*}
    \textbf{FID} = \| \vmu_0-\vmu\|^2_2\,+\,\mbox{tr}\left[\Sigma_0 + \Sigma - 2 \left(\Sigma_0 \Sigma\right)^{\frac{1}{2}} \right]
\end{align*}

We will need to privatize $\mu$ and $\Sigma$. Following \citet{park2017dp}, we privatize the mean vector using a $(\epsilon_1, \delta_1)$-DP Gaussian mechanism. Let us first recall the definition of $\vmu$ given by :
\begin{align*}
\vmu := \frac{1}{n}\sum_{i=1}^n \vphi(\vx_i).
\end{align*}
Assuming $\| \vphi(\vx_i) \|\leq 1$
for any $i$, the sensitivity of the mean vector denoted by $\Delta_{\vmu}$ is :
\begin{align*}
    \Delta_{\vmu}&=\max_{\vx_j, \vx_j'}\| \,\,\frac{1}{n}\big(\cancel{\sum_{i=1, i\neq j}^n \vphi(\vx_i)}+\vphi(\vx_j)\,\,\big)- \frac{1}{n}\big(\cancel{\sum_{i=1, i\neq j}^n \vphi(\vx_i)}+\vphi(\vx_j')\,\,\big)   \|\\
    &=\frac1n\max_{\vx_j, \vx_j'}\|\vphi(\vx_j)-\vphi(\vx_j')\|\\
    &\leq \frac1n \cdot 2\cdot \| \vphi(\vx_i) \|\\
    &=2/n
\end{align*}

i.e., bounded by $2/n$, when using the replacement definition of differential privacy (it is $1/n$ when using the inclusion/exclusion definition of DP).
Hence, 
\begin{align}\label{eq:mean}
\tilde\vmu := \vmu + \vn_1,
\end{align} where $\vn_1$ is drawn from $\Nrm(0, \Delta_\vmu^2 \sigma_1^2 I)$.
Here, $\sigma_1$ is a function of the privacy level given by $(\epsilon_1, \delta_1)$-DP. The exact relationship between $\sigma_1$ and $(\epsilon_1, \delta_1)$ varies depending on how to compute the DP bound. We use the bound introduced in the analytic Gaussian mechanism in \url{https://github.com/yuxiangw/autodp}.

Recall the definition of covariance given by:
\begin{align*}
    \Sigma = \frac1n X^T X-\vmu\vmu^T.
\end{align*} Since we have a privatized mean from above, we need to privatize the second-moment matrix $ \frac1n X^T X=M_{\text{sec}}$ to privatize the covariance matrix. 

As before, assuming $\| \vphi(\vx_i) \|\leq 1, \forall i$, the sensitivity of the second moment matrix denoted by $\Delta_{M_{\text{sec}}}$ is :
\begin{align*}
    \Delta_{M_{\text{sec}}} &= \max_{\vphi(\vx_j), \vphi(\vx_j')}\|\,\,\frac1n\big(\cancel{\sum_{i=1, i\neq j}^n \vphi(\vx_i)\vphi(\vx_i^T)}+\vphi(\vx_j)\vphi(\vx_j^T)\,\,\big)-\frac1n\big(\cancel{\sum_{i=1, i\neq j}^n \vphi(\vx_i)\vphi(\vx_i^T)}+\vphi(\vx_j')\vphi(\vx_j'^T)\,\,\big)\,\,\|_F\\
    &=\frac1n \max_{\vphi(\vx_j),\vphi(\vx_j')}\|\vphi(\vx_j)\vphi(\vx_j^T)-\vphi(\vx_j')\vphi(\vx_j'^T)\|_F\\
    &\leq \frac2n \max_{\vphi(\vx_j)}\|\vphi(\vx_j)\vphi(\vx_j^T)\|_F \hspace{2cm}(\text{WLOG, say }\|\vphi(\vx_j')\vphi(\vx_j'^T)\|_F\leq \|\vphi(\vx_j)\vphi(\vx_j^T)\|_F)\\
    &=\frac2n \max_{\vphi(\vx_j)}\sqrt{\sum_{i=1}^d\sum_{k=1}^d |\alpha_i\alpha_k|^2} \hspace{1.8cm}(\text{Say } \vphi(\vx_j)=[\alpha_1,\cdots,,\alpha_d] \text{ as a column vector})\\
    &=\frac2n \max_{\vphi(\vx_j)}\sqrt{\sum_{i=1}^d\big(\alpha_i^2\cdot\sum_{k=1}^d \alpha_k^2\big)}\\
    &=\frac2n \max_{\vphi(\vx_j)}\sqrt{\big(\sum_{i=1}^d\alpha_i^2\big)\cdot\big(\sum_{k=1}^d \alpha_k^2\big)}\\
    &=\frac2n \max_{\vphi(\vx_j)}\sqrt{\|\vphi(\vx_j)\|_2^2\cdot \|\vphi(\vx_j)\|_2^2}\\
    &\leq \frac2n
\end{align*}
i.e., bounded by $2/n$, when using the replacement definition of differential privacy (it is $1/n$ when using the inclusion/exclusion definition of DP). See eq(17) in \citet{park2017dp} for derivation of the sensitivity. 

Given a privacy budget $(\epsilon_2,\delta_2)$-DP assigned to this privatization step, which gives us a corresponding privacy parameter $\sigma_2$, we first draw noise $\vn_2$ from $\Nrm(0, \Delta_{M_{\text{sec}}}^2 \sigma_2^2 I_{d(d+1)/2})$ . We then add this noise to the upper triangular part of the matrix, including the diagonal component.
To ensure the symmetry of the perturbed second-moment matrix $\widetilde{M_{\text{sec}}} $, 
we copy the noise added to the upper triangular part and add one by one to the lower triangular part.
We then obtain 
\begin{align}\label{eq:cov}
    \tilde{\Sigma} = \widetilde{M_{\text{sec}}}-\tilde\vmu\tilde\vmu^T.
\end{align}
Note $\tilde{\Sigma}$ may not be positive definite. In such a case, we can project the negative eigenvalue to some small value close to zero to guarantee the positive definite property of the covariance matrix. This is safe to do, as DP is post-processing invariant. 

Using the aforementioned privatized mean given in \eqref{mean} and covariance given in \eqref{cov}, we can compute the final FID score, given by 
\begin{align*}
   \textbf{DP-FID} = \|\vmu_0 - \tilde{\vmu}\|_2^2 + \mbox{tr}\left[\Sigma_0 + \tilde{\Sigma} - 2 \left(\Sigma_0 \tilde{\Sigma}\right)^{\frac{1}{2}} \right],
\end{align*} which is 
 $(\epsilon_1+\epsilon_2, \delta_1+\delta_2)$-DP.

 Following the above method, we compute the DP-FID scores of SVHN, KMNIST, and EMNIST, with respect to private dataset MNIST. We sampled 60k data from each public dataset candidates to do a fair comparison, the results are in \tabref{dpfid} left. One could consider only privatizing the mean only, and the results are in \tabref{dpfid} right. Based on these results, we choose EMNIST as a public dataset.

 \begin{table}
\centering
\scalebox{0.8}{
\begin{tabular}{c|cccccc}
       $\epsilon$  & 0.1 & 0.5 & 1 & 2 & 5 & 10\\
       \hline
        SVHN & \textbf{78.9930} & \textbf{17.3465} & 8.9223 & 4.7426 & 2.1582 & 1.3368\\
        KMNIST & 79.9224 & 17.5794 & 9.0605 & 4.6092 & 1.9300 & 1.0794\\
        \textbf{EMNIST} & 80.4267 & 17.3617 & \textbf{8.9058} & \textbf{4.4561} & \textbf{1.8926} & \textbf{1.0242}\\
        \hline
    \end{tabular}
\quad
\begin{tabular}{c|cccc}
       $\epsilon$  & 0.1 & 0.5 & 1 & 2\\
       \hline
        SVHN & 0.2675 & 0.2662 & 0.2663 & 0.2663\\
        KMNIST & 0.0489 & 0.0459 & 0.0457 & 0.0457\\
        \textbf{EMNIST} & \textbf{0.0228} & \textbf{0.0201} & \textbf{0.0201} & \textbf{0.0197}\\
        \hline
    \end{tabular}}
\caption{
Left : Perturbed FIDs when privatizing both mean and covariance. Note $\epsilon$ listed is the sum of $\epsilon_{\mu}+\epsilon_{\Sigma}$, we take $\epsilon_{\mu}=\epsilon_{\Sigma}$. Right : Perturbed mean differences when privatizing mean only. $\epsilon$ listed is $\epsilon_{\mu}$}
\label{tab:dpfid}
\end{table}

\subsubsection{Additional experiments with SVHN and KMNIST} \label{supp:emnist-fid}

To verify our choice of EMNIST emperically, we also did ablation experiments on SVHN and KMNIST under the same privacy condition $\epsilon=10, \delta=10^{-5}$ to compare with EMNIST, the results are illustrated in \tabref{svhn-kmnist}.

\begin{table}[ht]
\centering
\scalebox{1.0}{
\begin{tabular}{llc}
&  Dataset pair   & CNN accuracy\\
 \hline
& (SVHN, MNIST) & 94.3  \\
& (KMNIST, MNIST)  & 96.3  \\
& \textbf{(EMNIST, MNIST)}  &  \textbf{97.4} \\
\hline
\end{tabular}}
\caption{
We also pretrained LDMs using SVHN and KMNIST then fine-tuned with MNIST, and list the CNN accuracy here respectively.}
\label{tab:svhn-kmnist}
\end{table}

\subsubsection{Ablation experiments for MNIST}\label{supp:emnist-abl}

\begin{wrapfigure}{r}{0.5\textwidth}
    \centering
    \vspace{-17mm}
    \includegraphics[width=0.3\textwidth]{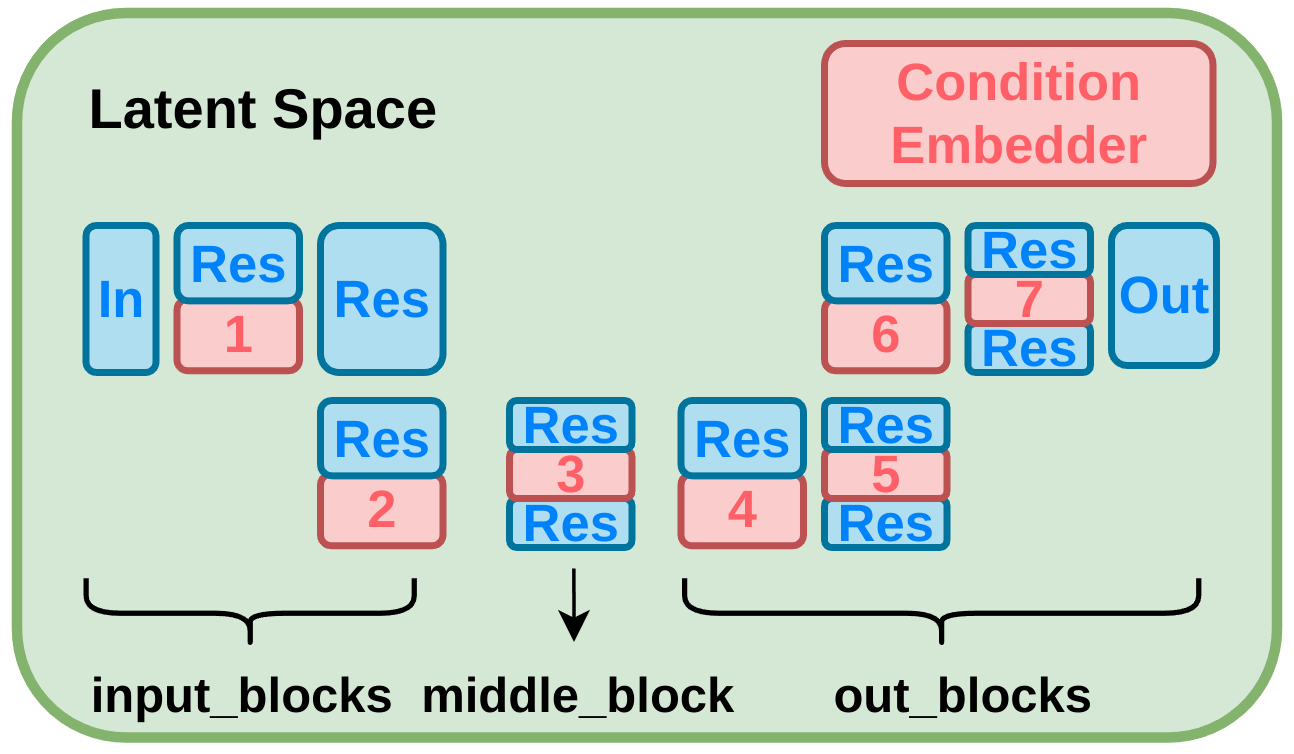}
    \caption{UNet Structure for MNIST.}
    \label{fig:mnist-unet}
\end{wrapfigure}

There are 7 attention modules in the UNet structure for MNIST, 1-2 are in input\_blocks, 3 is in middle\_block, 4-7 are in out\_blocks as illustrated in \figref{mnist-unet}. Modules in blue are frozen during fine-tuning. Parameters of condition embedder is always trained. We consider fine tune only $i$-th to 7th attention modules to reduce more trainable parameters. Results for $\epsilon=10, \delta=10^{-5}$ are listed in \tabref{emnist-ablation}. The best results is achieved when fine tune with 4-7 attention blocks, which means out\_blocks are more important than others during training.

\begin{table}[ht]

\centering
\scalebox{1.0}{
\begin{tabular}{llccccc}
&   &  1-7(all)  & 2-7 & 3-7 & 4-7 & 5-7\\
 \hline
& CNN & 97.3 & 97.3 & 90 & \textbf{97.4} & 97.3 \\
& \# of trainable params & 1.6M & 1.5M & 1.2M & 0.8M & 0.5M\\
& out of 4.6M total params & (34.3\%) & (32.4\%) & (25.2\%) & (18.0\%) & (10.9\%)\\
\hline
\end{tabular}}
\caption{
CNN accuracy and number of trainable parameters for MNIST ablation experiments with varying number of fine-tuning layers. Privacy condition is set to $\epsilon=10, \delta=10^{-5}$.}
\label{tab:emnist-ablation}
\end{table}

\subsection{Transferring from Imagenet to CIFAR10 distribution}\label{supp:imagenet_to_cifar}

\begin{figure}
    \centering
    \includegraphics[width=0.9\textwidth]{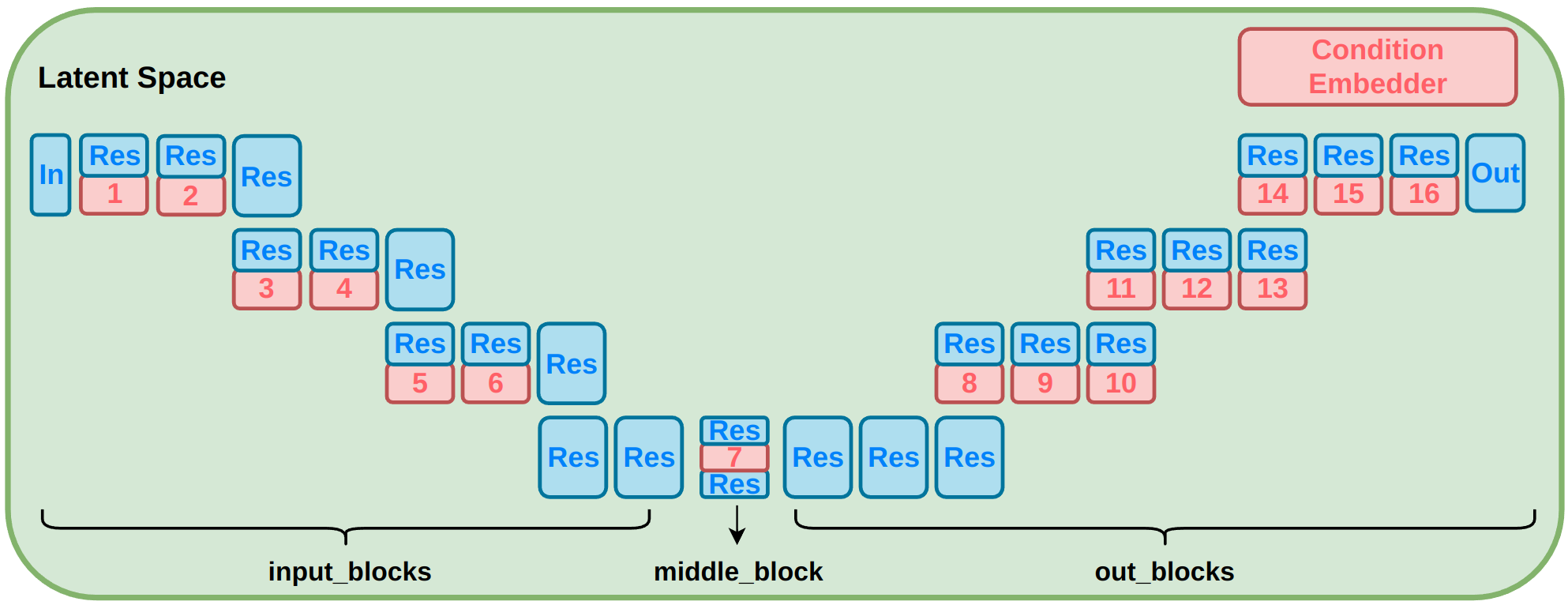}
    \caption{UNet Structure for CIFAR-10}
    \label{fig:cifar-unet}
\end{figure}

Here, we provide the results for ablation experiments to test the performance of DP-LDM when fine-tuning only certain attention modules inside the pre-trained model and keeping the rest of the parameters fixed.
There are 16 attention modules in total as illustrated in \figref{cifar-unet}.
\tabref{cifar_finetuning_layers} shows the FID obtained for $\epsilon = 1, 5, 10$ and $\delta=10^{-5}$ for the different number of attention modules fine-tuned. The results show that fixing up to the first half of the attention layers in the LDM has a positive effect in terms of the FID (the lower the better) in our model.

\begin{table}[ht]
\centering
\scalebox{1}{
\begin{tabular}{lccc}
   DP-LDM        &  $\epsilon=10$ & $\epsilon=5$ & $\epsilon=1$  \\
 \hline
 1-16 layers & 25.8 $\pm$ 0.3  & 29.9 $\pm$ 0.2 & 33.0 $\pm$ 0.3 \\
5 - 16 layers & 15.7 $\pm$ 0.3   & 21.2  $\pm$ 0.2 &  28.9 $\pm$ 0.2   \\
9 - 16 layers  &  \textbf{8.4 $\pm$ 0.2}   & \textbf{13.4 $\pm$ 0.4} &   \textbf{22.9 $\pm$ 0.5}  \\
13 - 16  layers &  12.3 $\pm$ 0.2 &   18.5 $\pm$ 0.2  &   25.2 $\pm$  0.5\\
\hline
\end{tabular}}
\caption{FID scores (lower is better) for synthetic CIFAR-10 data with varying the number of fine-tuning layers and privacy guarantees. \textbf{Top row (1-16 layers):} Fine-tuning all attention modules. \textbf{Second row (5-16 layers):} Keep first 4 attention modules fixed and fine-tuning from 5 to 16 attention modules. \textbf{Third row (9-16 layers):}  Keep first 8 attention modules fixed and fine-tuning from 9 to 16 attention modules. \textbf{Bottom row (13-16 layers):}  Keep first 12 attention modules fixed and fine-tuning from 13 to 16 attention modules.}
\label{tab:cifar_finetuning_layers}
\end{table}

We also report the different hyper-parameter settings used in ablation experiments in table \tabref{cifar10_dp_hyperparam}.

\begin{table*}[ht]
\centering
\scalebox{0.9}{
\begin{tabular}{ll  ccc}
\toprule
& &  $\epsilon=10$ & $\epsilon=5$ & $\epsilon=1$\\
\midrule
\multirow{4}{*}{1-16 layers ($24.4$M parameters)} 
& batch size & 1000 & 2000 & 1000\\
& clipping norm & $10 ^{-5}$ & $10 ^{-5}$ & $10 ^{-3}$\\
& learning rate & $10 ^{-6}$& $10 ^{-6}$ & $10 ^{-6}$\\
& epochs & 30 & 30 & 10\\
\midrule
 \multirow{4}{*}{5-16 layers ($20.8$M parameters)} 
 & batch size & 5000 & 5000 & 2000\\
& clipping norm & $10 ^{-6}$ & $10 ^{-5}$ & $10 ^{-3}$\\
& learning rate & $10 ^{-6}$ & $10 ^{-6}$ & $10 ^{-5}$\\
& epochs & 50 & 50 & 10\\
\midrule
 \multirow{4}{*}{9-16 layers ($10.2$M parameters)} 
  & batch size &  2000 & 2000 & 5000\\
 & clipping norm  & $10 ^{-6}$ & $10 ^{-6}$ & $10 ^{-2}$\\
& learning rate & $10 ^{-6}$ & $10 ^{-6}$ & $10 ^{-6}$\\
& epochs & 30 & 30 & 10\\
\midrule
 \multirow{4}{*}{13-16 layers ($4$M parameters)} 
   & batch size & 2000 & 2000 & 2000\\
 & clipping norm & $10 ^{-6}$ & $10 ^{-6}$ & $10 ^{-2}$\\
& learning rate & $10 ^{-6}$ & $10 ^{-6}$ & $10 ^{-6}$ \\
& epochs & 30 & 30 & 10\\
\bottomrule
\end{tabular}}
\caption{
DP-LDM hyper-parameter setting on CIFAR-10 for different ablation experiments.}
\label{tab:cifar10_dp_hyperparam}
\end{table*}

\tabref{cifar10_wrn_resnet_hyper} shows the hyper-parameters used during training ResNet9 and WRN-40-4 downstream classifiers on CIFAR10 synthetic samples. 

\begin{table}[tb]
\centering
\scalebox{1.0}{
\begin{tabular}{lcc}
& ResNet9 & WRN-40-4 \\
\hline
    learning rate & 0.5  & 0.1   \\
    batch size  & 512  & 1000       \\
    epochs           & 10          & 10000            \\
    optimizer            & SGD           & SGD      \\
    label smoothing & 0.1 &  0.0 \\
    weight decay &  $5 \cdot 10 ^{-4}$ & $5 \cdot 10 ^{-4}$ \\
    momentum &  0.9 & 0.9 \\
\hline
\end{tabular}}
\caption{
Hyperparameters for downstream classification ResNet9 and WRN-40-4 trained on CIFAR10 synthetic data}
\label{tab:cifar10_wrn_resnet_hyper} 
\end{table}

\subsection{Transferring from Imagenet to CelebA32}
\label{supp:imagenet_to_celeba32}

\begin{figure}[t]
\centering
\includegraphics[width=0.8\textwidth]{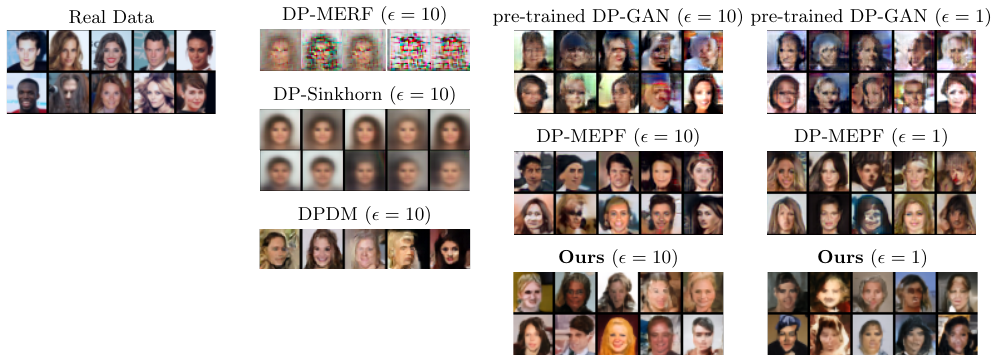}
\caption{
 Synthetic $32 \times 32$ CelebA samples generated at different levels of privacy. Samples for DP-MERF and DP-Sinkhorn are taken from \citep{dp_sinkhorn}, DPDM samples are taken from \citep{nvidia_DPDM}, and DP-MEPF samples are taken from \citep{DP-MEPF}.}
\label{fig:celeba32_samples}
\end{figure}

We also apply our model in the task of generating $32\times32$ CelebA images. The same pretrained autoencoder as our CIFAR-10 experiments in Section \ref{sec:simpletransfer} is used, but since this experiment is for unconditional generation, we are unable to re-use the LDM. A new LDM is pretrained on Imagenet without class conditioning information, and then fine-tuned on CelebA images scaled and cropped to $32\times32$ resolution. Our FID results for $\delta=10^{-6}$, $\epsilon=1, 5, 10$ are summarized in Table \ref{tab:celeba32_fid}. We achieve similar results to DP-MEPF for $\epsilon=5$ and $\epsilon=10$. As with our results at $64\times64$ resolution, our LDM model does not perform as well in higher privacy settings ($\epsilon=1$). Sample images are provided in Fig. \ref{fig:celeba32_samples}

\begin{table}[tb]
\centering
\scalebox{1.0}{
\begin{tabular}{lccc}
& $\epsilon=10$ & $\epsilon=5$ & $\epsilon=1$ \\
\hline
    \textbf{DP-LDM} (ours, average) & 16.2 $\pm$ 0.2 & 16.8 $\pm$ 0.3 & 25.8 $\pm$ 0.9 \\
    \textbf{DP-LDM} (ours, best)    & \textbf{16.1}  & \textbf{16.6}  & 24.6           \\
    DP-MEPF $(\vphi_1)$             & 16.3           & 17.2           & \textbf{17.2}  \\
    DP-GAN (pre-trained)            & 58.1           & 66.9           & 81.3           \\
    DPDM (no public data)           & 21.2           &  -             & 71.8           \\
    DP Sinkhorn (no public data)    & 189.5          &  -             &  -             \\
\hline
\end{tabular}
}
\caption{
CelebA FID scores (lower is better) for images of resolution $32 \times 32$ comparing with results from DPDM \citep{nvidia_DPDM}, DP Sinkhorn \citep{dp_sinkhorn}, and DP-MEPF \citep{DP-MEPF}.}
\label{tab:celeba32_fid}
\end{table}

\subsection{Transferring from Imagenet to Camelyon17-WILDS}
\label{supp:imagenet_to_camelyon}

Camelyon17-WILDS is part of the WILDS benchmark suite of datasets, containing $455,954$ medical images at $96 \times 96$ resolution. The downstream task is to determine whether the center $32 \times 32$ patch of the image contains any tumor pixels. In our experiment, we crop the image so that only the center $32 \times 32$ patch is passed to the model. We begin with the same pretrained autoencoder and LDM as in our CIFAR-10 experiments (Section \ref{sec:simpletransfer}), and fine-tune on Camelyon17-WILDS. We then generate $25,000$ images conditioned on each of the two classes, and combine them to create a synthetic dataset of $50,000$ images. Finally, the synthetic dataset is used to train a WRN-40-4 classifier. 

\section{Hyperparameters}\label{supp:hyper}

Here we provide an overview of the hyperparameters of the pretrained autoencoder in \tabref{autoencoder}, hyperparameters of the pretrained diffusion models in \tabref{dm}. Note that $base\  learning\ rate$ is the one set in the yaml files. The real learning rate passed to the optimizer is $accumulate\_grad\_batches$ $\times$ $num\_gpus$ $\times$ $batch\ size$ $\times$ $base\ learning\ rate$.

\begin{table*}[h]
    \centering
    \scalebox{0.8}{
    \begin{tabular}{l c c c c}
    \toprule
                            &  EMNIST     & ImageNet     & ImageNet  & ImageNet \\
                            &  (to MNIST) & (to CIFAR10) & (to CelebA 32) & (to CelebA 64)\\
    \midrule
      Input size            & 32          & 32           & 32          & 64 \\
      Latent size           & 4           & 16           & 16          & 32 \\
      $f$                   & 8           & 2            & 2           & 2 \\
      $z$-shape & $4\times4\times3$  & $16\times16\times3$ & $16\times16\times3$ & $64\times64\times3$ \\
      Channels              & 128         & 128          & 128         & 192 \\
      Channel multiplier    & [1,2,3,5]   & [1,2]        & [1,2]       & [1,2] \\
      Attention resolutions & [32,16,8]   & [16, 8]      & [16, 8]     & [16,8] \\
      num\_res\_blocks      & 2           & 2            & 2           & 2 \\
      Batch size            & 50          & 16           & 16          & 16 \\
      Base learning rate & $4.5\times 10^{-6}$ & $4.5\times 10^{-6}$ & $4.5\times 10^{-6}$ & $1.0\times10^{-6}$ \\
      Learning rate & $4.5\times 10^{-4}$  & $1.4\times10^{-4}$ & $1.4\times10^{-4}$ & $1.4\times10^{-4}$ \\
      Epochs                & 50          & 2            & 2   & - \\
      GPU(s) & 1 NVIDIA V100 & 1 NVIDIA RTX A4000 & 1 NVIDIA RTX A4000 & 1 NVIDIA V100 \\
      Time                  & 8 hours     &  1  day      & 1 day      & 1 day \\
      \bottomrule
    \end{tabular}}
    \caption{
    Hyperparameters for the pretrained autoencoders for different datasets. }
    \label{tab:autoencoder}
\end{table*}

\begin{table*}[h]
    \centering
    \scalebox{0.8}{
    \begin{tabular}{l c c c c c}
    \toprule
                                  &  EMNIST      & ImageNet     & ImageNet       & ImageNet \\
                                  &  (to MNIST)  & (to CIFAR10) & (to CelebA 32) & (to CelebA64) \\
    \midrule
        input size                & 32           & 32           & 32             & 64 \\
        latent size               & 4            & 16           & 16             & 32 \\
        $f$                       & 8            &  2           & 2              & 2 \\
        $z$-shape & $4\times4\times3$ & $16\times16\times3$ & $16\times16\times3$ & $32\times32\times3$ \\
        channels                  & 64           &  128         & 192            & 192 \\
        channel multiplier        & [1,2]        & [1,2,2,4]    & [1,2,4]      & [1,2,4] \\
        attention resolutions     & [1,2]        & [1,2,4]      & [1,2,4]      & [1,2,4] \\
        num\_res\_blocks          & 1            & 2            & 2              & 2 \\
        num\_heads                & 2            & 8            & -              & 8 \\
        num\_head\_channels       & -            & -            & 32             & - \\
        batch size                & 512          & 500          & 384            & 256 \\
        base learning rate & $5\times10^{-6}$ & $1\times10^{-6}$ & $5\times10^{-7}$ & $1\times10^{-6}$ \\
        learning rate & $2.6\times10^{-3}$ & $5\times10^{-4}$ & $2\times10^{-4}$  & $2.6\times10^{-4}$ \\
        epochs                    & 120           & 30           & 13             & 40 \\
        \# trainable parameters   & 4.6M         & 90.8M        & 162.3M         & 72.2M \\
        GPU(s) & 1 NVIDIA V100 &  1 NVIDIA RTX A4000 & 1 NVIDIA V100 & 1 NVIDIA V100 \\
        time  &   6 hours     &  7 days & 30 hours     & 10 days \\
        \midrule
        use\_spatial\_transformer & True         & True         & False          & False \\
        cond\_stage\_key          & class\_label & class\_label & -              & - \\
        conditioning\_key         & crossattn    & crossattn    & -              & - \\
        num\_classes              & 26           & 1000         & -              & - \\
        embedding dimension       & 5            & 512          & -              & - \\
        transformer depth         & 1            & 1            & -              & - \\
        \bottomrule
    \end{tabular}}
    \caption{
    Hyperparameters for the pretrained diffusion models for different datasets.}
    \label{tab:dm}
\end{table*}

\tabref{finetune-mnist} shows the hyperparameters we used for fine-tuning on MNIST. \tabref{finetune-celeba32} shows the hyperparameters we used for CelebA32. \tabref{finetune-celeba64} shows the hyperparameters we used for CelebA64. \tabref{finetune-txt2img} shows the hyperparameters we used for text-to-image CelebAHQ generation. \tabref{finetune-celebahq} shows the hyperparmeters we used for class-conditioned CelebAHQ generation.

\begin{table*}[h]
    \centering
    \scalebox{0.8}{
    \begin{tabular}{l c c }
    \toprule
         &  $\epsilon=10$ & $\epsilon=1$ \\
    \midrule
      batch size & 2000  &  2000 \\
      base learning rate & $5\times 10^{-7}$ & $6\times 10^{-7}$\\
      learning rate & $1\times 10^{-3}$ &  $1.2\times 10^{-3}$ \\
      epochs & 200  & 200 \\
      clipping norm & 0.01 & 0.001 \\
      noise scale & 1.47 &   9.78\\
      ablation & 4 & -1 \\
      num of params & 0.8M & 1.6M \\
      \midrule
      use\_spatial\_transformer & True & True\\
      cond\_stage\_key &  class\_label & class\_label\\
      conditioning\_key & crossattn & crossattn\\
      num\_classes &  26  &  26 \\
      embedding dimension & 13 &  13\\
      transformer depth & 1 &  1 \\
      train\_condition\_only & True & True\\
      attention\_flag &  spatial & spatial\\
      \# condition params & 338 &  338\\
      \bottomrule
    \end{tabular}}
    \caption{
    Hyperparameters for fine-tuning diffusion models with DP constraints $\epsilon=10,1$ and $\delta=10^{-5}$ on MNIST. The ``ablation'' hyperparameter determines which attention modules are fine-tuned, where a value of $i$ means that the first $i-1$ attention modules are frozen and others are trained. Setting ``ablation'' to $-1$ (default) fine-tunes all attention modules.}
    \label{tab:finetune-mnist}
\end{table*}

\begin{table*}[h]
    \centering
    \scalebox{0.8}{
    \begin{tabular}{l c c c}
    \toprule
         &  $\epsilon=10$ & $\epsilon=5$ & $\epsilon=1$ \\
    \midrule
      batch size & 8192 &  8192 & 2048 \\
      base learning rate & $5\times10^{-7}$ & $5\times10^{-7}$ & $5\times10^{-7}$\\
      learning rate & $4\times 10^{-3}$ &  $4\times 10^{-3}$ & $1\times 10^{-3}$ \\
      epochs & 20 & 20 & 20\\
      clipping norm & $5.0\times10^{-4}$ & $5.0\times10^{-4}$ & $5.0\times10^{-4}$ \\
      ablation & -1 & -1 & -1 \\
      \midrule
      use\_spatial\_transformer & False & False & False \\
      cond\_stage\_key & - & - & - \\
      conditioning\_key & - & - & - \\
      num\_classes & - & - & - \\
      embedding dimension & - & - & -\\
      transformer depth & - & - & - \\
      train\_attention\_only & True & True & True \\
      \bottomrule
    \end{tabular}}
    \caption{
    Hyperparameters for fine-tuning diffusion models with DP constraints $\epsilon=10,5,1$ and $\delta=10^{-6}$ on CelebA32.}
    \label{tab:finetune-celeba32}
\end{table*}

\begin{table*}[h]
    \centering
    \scalebox{0.8}{
    \begin{tabular}{l c c c}
    \toprule
         &  $\epsilon=10$ & $\epsilon=5$ & $\epsilon=1$ \\
    \midrule
      batch size & 8192 &  8192 & 8192\\
      base learning rate & $1\times10^{-7}$ & $1\times10^{-7}$ & $1\times10^{-7}$\\
      learning rate & $8.2\times 10^{-4}$ &  $8.2\times 10^{-4}$ & $8.2\times 10^{-4}$ \\
      epochs & 70 & 70 & 70\\
      clipping norm & $5.0\times10^{-4}$ & $5.0\times10^{-4}$ & $5.0\times10^{-4}$ \\
      ablation & -1 & -1 & -1 \\
      \midrule
      use\_spatial\_transformer & False & False& False\\
      cond\_stage\_key &  - & - & - \\
      conditioning\_key & - & - & - \\
      num\_classes & - & - & - \\
      embedding dimension & - & - & - \\
      transformer depth & - & - & - \\
      train\_attention\_only & True & True & True \\
      \bottomrule
    \end{tabular}}
    \caption{
    Hyperparameters for fine-tuning diffusion models with DP constraints $\epsilon=10,5,1$ and $\delta=10^{-6}$ on CelebA64.}
    \label{tab:finetune-celeba64}
\end{table*}

\begin{table*}[h]
    \centering
    \scalebox{0.8}{
    \begin{tabular}{l c}
    \toprule
         &  $\epsilon=10$  \\
    \midrule
      batch size & 20000 \\
      base learning rate & $1.0\times10^{-7}$ \\
      learning rate & $2.0\times 10^{-3}$ \\
      epochs & 30 \\
      clipping norm & $1.0\times10^{-6}$ \\
      ablation & 13 \\
      \midrule
      use\_spatial\_transformer & True \\
      cond\_stage\_key & class\_label \\
      conditioning\_key & crossattn \\
      num\_classes & 1001 \\
      embedding dimension & 512 \\
      transformer depth & 1 \\
      train\_condition\_only & True \\
      attention\_flag & spatial \\
      \# condition params & $512,512$ \\
      \bottomrule
    \end{tabular}}
    \caption{
    Hyperparameters for fine-tuning diffusion models with DP constraint $\epsilon=10$ and $\delta=3\times10^{-6}$ on Camelyon17-WILDS.}
    \label{tab:finetune-camelyon}
\end{table*}

\begin{table*}[h]
    \centering
    \scalebox{0.8}{
    \begin{tabular}{l c c }
    \toprule
         &  $\epsilon=10$ & $\epsilon=1$ \\
    \midrule
      batch size & 256  &  256 \\
      base learning rate & $1\times 10^{-7}$ & $1\times 10^{-7}$\\
      learning rate & $2.6\times 10^{-5}$ &  $2.6\times 10^{-5}$ \\
      epochs & 10  & 10 \\
      clipping norm & 0.01 & 0.01 \\
      noise scale & 0.55 &   1.46\\
      ablation & -1 & -1 \\
      num of params & 280M & 280M \\
      \midrule
      use\_spatial\_transformer & True & True\\
      cond\_stage\_key &  caption & caption\\
      context\_dim &  1280  & 1280 \\
      conditioning\_key & crossattn & crossattn\\
      transformer depth & 1 &  1 \\
      \bottomrule
    \end{tabular}}
    \caption{
    Hyperparameters for fine-tuning diffusion models with DP constraints $\epsilon=10,1$ and $\delta=10^{-5}$ on text-conditioned CelebAHQ.}
    \label{tab:finetune-txt2img}
\end{table*}

\begin{table*}[h]
    \centering
    \scalebox{0.8}{
    \begin{tabular}{l c c c}
    \toprule
         &  $\epsilon=10$ & $\epsilon=5$ & $\epsilon=1$ \\
    \midrule
      batch size & 2048 & 2048 & 2048 \\
      base learning rate & $1\times 10^{-7}$ & $1\times 10^{-7}$ & $1\times 10^{-7}$ \\
      learning rate & $2.0\times 10^{-4}$ & $2.0\times 10^{-4}$ & $2.0\times 10^{-4}$ \\
      epochs & 50  & 50 & 50 \\
      clipping norm & $5.0\times10^{-4}$ & $5.0\times10^{-4}$ & $5.0\times10^{-4}$ \\
      ablation & -1 & -1 & -1 \\
      \midrule
      use\_spatial\_transformer & True & True & True\\
      cond\_stage\_key & class\_label & class\_label & class\_label\\
      context\_dim &  512  & 512 & 512 \\
      conditioning\_key & crossattn & crossattn & crossattn\\
      transformer depth & 1 & 1 & 1 \\
      \bottomrule
    \end{tabular}}
    \caption{
    Hyperparameters for fine-tuning diffusion models with DP constraints $\epsilon=10,5,1$ and $\delta=10^{-5}$ on class-conditional CelebAHQ.}
    \label{tab:finetune-celebahq}
\end{table*}

\FloatBarrier
\section{Additional Samples}

\begin{figure}[h]
\centering
\includegraphics[width=\textwidth]{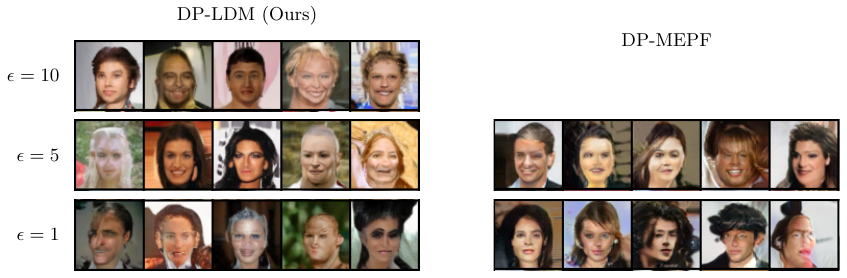}
\caption{
 Synthetic $64 \times 64$ CelebA samples generated at different levels of privacy. Samples for DP-MEPF are taken from \citet{DP-MEPF}.}
\label{fig:celeba64_samples}
\end{figure}

\begin{figure}[h]
\centering
\begin{tabular}{cc}
  DP-LDM (Ours)  &  DP-Diffusion \\
  \includegraphics{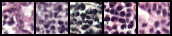} & \includegraphics[trim=1 98 288 31,clip]{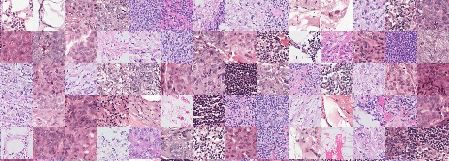} \\
  \includegraphics{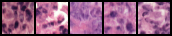} & \includegraphics[trim=1 66 288 63,clip]{images/camelyon/dp-diffusion/figure-1-camelyon-synthetic.jpg} \\
\end{tabular}
\caption{
 Synthetic $32 \times 32$ Camelyon17-WILDS samples at $\epsilon=10$ and $\delta=3\times10^{-6}$. Samples for DP-Diffusion are taken from table 1 in \citet{deepmind_DPDM}.}
\label{fig:camelyon_samples}
\end{figure}

\FloatBarrier
\section{Comparison between DP and non-DP Samples}

\begin{figure}[h]
\centering
\includegraphics[scale=0.4]{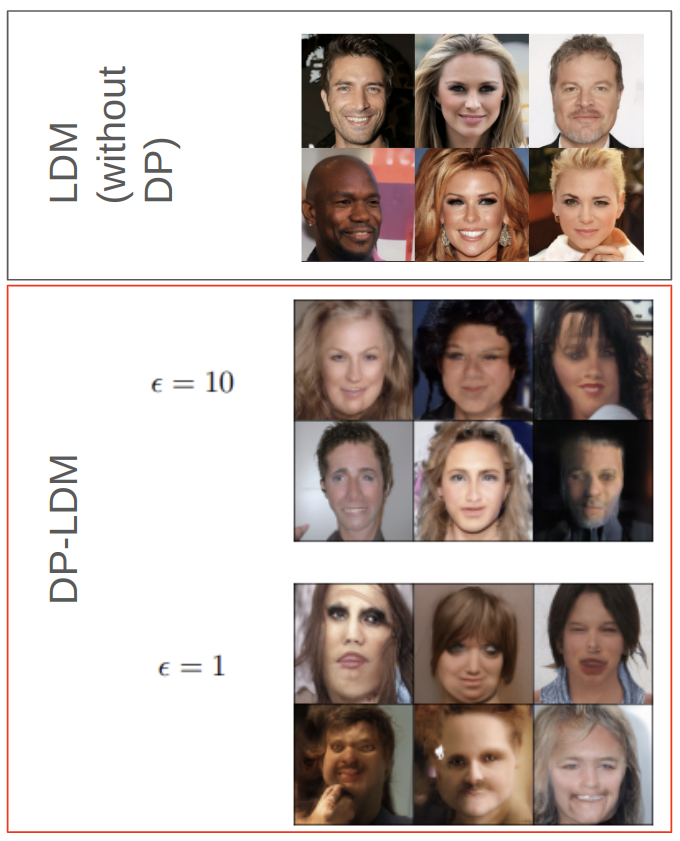}
\caption{
 Synthetic CelebAHQ images from a non-DP-trained LDM (Top, black box, images taken from \citet{LDM}) versus those from DP-trained LDMs (Bottom, red box). The non-DP synthetic images exhibit humanly discernible characteristics of faces. On the other hand, DP synthetic samples exhibit distortions at varying levels, where a higher privacy guarantee ($\epsilon=1$) yields more distortion.
 }
\label{fig:dp_nondp_chq}
\end{figure}

\end{document}